\documentclass[letterpaper]{article}
\usepackage[preprint]{paperstyle}
\usepackage[hyphens]{url}
\usepackage{graphicx}
\usepackage{natbib}
\usepackage{caption}
\usepackage{algorithm}
\usepackage{algorithmic}

\usepackage{newfloat}
\usepackage{listings}
\DeclareCaptionStyle{ruled}{labelfont=normalfont,labelsep=colon,strut=off}
\floatstyle{ruled}
\newfloat{listing}{tb}{lst}{}
\floatname{listing}{Listing}

\usepackage{booktabs}
\usepackage{multirow}
\usepackage{subcaption}

\usepackage{amsmath}
\usepackage{amssymb}

\newcommand{\bA}{\boldsymbol{A}}
\newcommand{\bv}{\boldsymbol{v}}

\newcommand{\bt}{\boldsymbol{t}}

\newcommand{\bz}{\boldsymbol{z}}

\newcommand{\rsrc}{\boldsymbol{r}^{\mathrm{src}}}
\newcommand{\rilr}{\boldsymbol{r}^{\mathrm{ilr}}}
\newcommand{\rsrcH}[1]{\boldsymbol{r}^{\mathrm{src}}_{#1}}

\title{Role-Break in Attention Heads:\\Understanding and Detecting Hallucinations in VLMs}
\author{
    Mingyu Wang\textsuperscript{\rm 1},
    Weilin Jin\textsuperscript{\rm 1},
    Wenbo Li\textsuperscript{\rm 2},
    Haoyang Huang\textsuperscript{\rm 2},
    Nan Duan\textsuperscript{\rm 2},
    Tong Jia\textsuperscript{\rm 1}\thanks{Corresponding author.},
    Chaoran Luo\textsuperscript{\rm 1},
    Ying Li\textsuperscript{\rm 1}\footnotemark[1]
}
\affiliations{
    \textsuperscript{\rm 1}Peking University, Beijing, China\\
    \textsuperscript{\rm 2}Joy Future Academy, Beijing, China
}

\begin{document}

\maketitle

\begin{abstract}
Despite remarkable progress in vision-language generation, Vision-Language Models (VLMs) remain prone to hallucinations, producing content that is inconsistent with or unsupported by the input image. Existing works largely design detection or mitigation methods around one specific hallucination pattern, such as visual-textual imbalance, but real VLM hallucinations arise from a mixture of multiple patterns, so signals bound to a single pattern struggle to remain stable across models and tasks. Under a unified head-level view, we find that hallucination-induced changes manifest as localized deviations from each head's faithful contextual behavior, a phenomenon we term Role-Break. Detailed analysis reveals that these deviations are systematically organized across attention heads, contextual sources, and deviation directions, and that the resulting signal is linearly readable once head identity is preserved. Based on these findings, we build a lightweight linear detector on top of Role-Break that requires no fine-tuning of the VLM, whose feature dimension stays below $5{,}000$ and reaches an average AUROC of $93.23$ across six VLMs and four benchmarks. A small-scale intervention experiment further shows that the detected tokens can be directly acted upon in the discriminative setting.
\end{abstract}

\section{Introduction}

Vision-language models (VLMs) have recently achieved substantial progress on image captioning, visual question answering, and multimodal reasoning. However, these models remain prone to generating content that is unsupported by the input image yet linguistically plausible, a phenomenon commonly referred to as hallucination. Hallucination limits the reliability and deployability of VLMs. Understanding and detecting hallucination has therefore become a central problem in multimodal model research.

\begin{figure}[t]
    \centering
    \includegraphics[width=\linewidth]{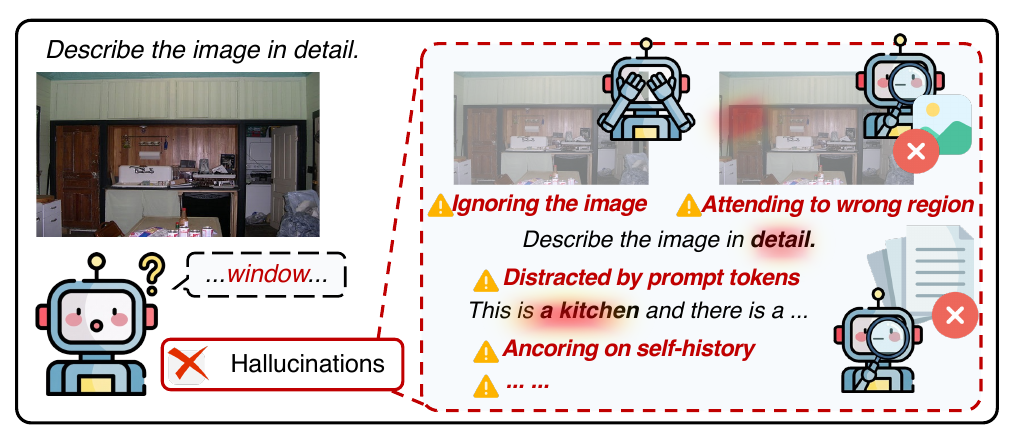}
    \caption{VLM hallucinations arise from multiple distinct patterns rather than a single mechanism.}
    \label{fig:intro_modes}
\end{figure}

Prior work studies hallucination from a variety of perspectives, commonly attributing it to behaviors such as weakened visual grounding, dominance of language priors, or abnormal cross-modal interaction, and further deriving hallucination indicators from output statistics, visual attention, or hidden states~\cite{huang2024opera,liu2024paying,leng2024mitigating,jiang2025devils,zhang2025dhcp}. Most of these methods design a detection or mitigation strategy around a specific observed phenomenon. Our key observation, however, is that real hallucinations arise from a mixture of patterns (Figure~\ref{fig:intro_modes}). Within a single model, hallucinations already split into several reproducible patterns whose organization further shifts across tasks, making signals bound to a single pattern struggle to remain stable across settings.

Concretely, we analyze VLM hallucination under a unified head-level view and reach three core findings. First, when the mechanisms behind existing accounts are cast into simplified head-level attention signatures and evaluated under a common protocol, no single signature stands out as a cross-setting-stable hallucination indicator. Second, every attention head exhibits a stable and highly heterogeneous behavior pattern on faithful tokens, which we term its faithful role. On hallucinated tokens, individual heads exhibit reproducible, systematic departures from their own roles, which we name Role-Break. Third, the predictive information carried by Role-Break is linearly readable.

Based on these findings, we use Role-Break as a hallucination detection signal. A simple linear classifier based on Role-Break reaches an average AUROC of $93.23$ across six VLMs and four benchmarks. The method is notably efficient and practical. On one hand, it requires no additional training of the VLM and introduces no extra VLM inference steps. On the other hand, our detector has a very simple structure, a linear classifier over the flattened per-head deviation vector, whose input dimensionality stays below $5{,}000$ on all six VLMs, so the inference-time compute overhead is negligible.

\paragraph{Contributions.}
\begin{itemize}
    \item We propose faithful role and Role-Break, characterizing VLM hallucination as a structured departure of multiple attention heads from their per-head faithful baselines.
    \item We reveal the necessity, the internal structure, and the appropriate representation of Role-Break.
    \item We propose a simple linear hallucination detector based on Role-Break, reaching an average AUROC of $93.23$ across six VLMs and four benchmarks, without any extra VLM training or additional VLM inference steps.
\end{itemize}

\section{Related Work}

\paragraph{Hallucinations in VLMs.}
VLMs integrate visual encoders with LLMs through projection layers to enable multimodal reasoning. VLM hallucinations mainly manifest as generated content deviating from the visual evidence in the input image, and are commonly grouped into object, attribute, and relational hallucinations~\cite{li2023evaluating, wang2023amber}. Existing evaluation protocols fall into two categories, discriminative~\cite{li2023evaluating,wang2023amber} and generative~\cite{rohrbach2018object,gunjal2024detecting}.

\paragraph{Hallucination Detection and Mitigation.}
Existing detection and mitigation methods start from different perspectives on hallucination. Some take specific behavioral pathologies as the starting point, such as over-reliance on anchor tokens~\cite{huang2024opera}, imbalance between visual and textual modalities~\cite{liu2024paying,zhao2025tell,tu2026attention}, or language-prior dominance~\cite{leng2024mitigating}. Others ground their design in particular attention heads or internal representations that carry the hallucination signal, such as insufficient mid-layer visual attention~\cite{jiang2025devils}, a set of hallucination-sensitive heads whose outputs carry the signal~\cite{zhang2026vib}, cross-modal attention weights~\cite{zhang2025dhcp}, or a fixed set of token-level statistics~\cite{fieback2024metatoken}. These works collectively show that hallucination manifests in diverse internal signals of VLMs.

\paragraph{Head-Level Analysis of Hallucination in VLMs.}
Some works analyze VLM hallucination at the level of individual attention heads and design head-targeted interventions. Most identify a subset of heads relevant to hallucination and act on them, such as scoring visually sensitive heads~\cite{he2024cracking,ji2026causallens,ibn2025fixing}, propagating shallow-layer visual sinks~\cite{zhang2024seeing,zhang2025shallow}, aligning head attention with the information flow~\cite{zhao2025mitigating,xu2026mitigating}, or identifying hallucination-inducing heads via causal mediation~\cite{sarkar2025mitigatinghallucinationsvisionlanguagemodels}. Others approach the problem from angles such as prompt-imposed object counts~\cite{rudman2026mechanisms}, instruction-token hijacking of visual saliency~\cite{chen2025attention}, or knowledge stability across layers~\cite{li2026saked}. Mechanistically, however, most of these methods design around one specific head-level pattern.

\section{Understanding Role-Break in VLMs}
\label{sec:understanding}

Attention heads in a trained VLM tend toward functional specialization. On faithful samples, different heads take on differentiated information-interaction tasks~\cite{olsson2022context,wang2022interpretability}, so each head develops its own characteristic behavior pattern. This head-level view is not bound to any single hallucination pattern and can therefore accommodate the mixed hallucination modes observed in practice. In this section, we analyze VLM hallucination from this head-level perspective along three angles: the head-dependent nature of attention signals, the definition and structure of Role-Break, and its linear readability as a predictive signal.

\subsection{Preliminary}
\label{sec:preliminary}

\paragraph{Task formulation.}
We study token-level hallucination detection in vision-language generation. A VLM receives an image and a text prompt, then autoregressively generates a caption $y_1, y_2, \ldots, y_T$. A token $y_k$ is labeled \emph{hallucinated} if it refers to content (an object, attribute, or relation) that is not supported by the image and \emph{faithful} otherwise. The goal is to decide, from the model's internal state at step $k$, whether $y_k$ is faithful or hallucinated.

\paragraph{Head-level source-allocation.}
The VLM's decoder is a Transformer with $L$ layers and $H$ heads. At layer $l$, head $h$, and decode step $k$, let $\bA_{k,l,h}$ denote the attention weights from the decoded token $y_k$ to the context tokens available to it. Each context token belongs to one of four disjoint groups: system-prompt, image, user-text, or self-generated. For each group $s \in \{\mathrm{sys}, \mathrm{img}, \mathrm{txt}, \mathrm{slf}\}$, we aggregate the attention mass that head $(l,h)$ places on that group:
\begin{equation}
    \pi_{k,l,h,s} \;\triangleq\; \sum_{i \in \mathcal{I}_s} \bA_{k,l,h}[i],
    \label{eq:vpartition}
\end{equation}
where $\mathcal{I}_s$ indexes group $s$. Stacking the four values yields a source-allocation vector $\boldsymbol{\pi}_{k,l,h}$ that summarizes how head $(l,h)$ divides its outgoing attention mass at token $y_k$, and its collection across heads is denoted $\boldsymbol{\pi}_k \in \mathbb{R}^{4LH}$.

\paragraph{Setup.}
Following prior work~\cite{jiang2025devils,zhang2026vib}, we study two VLMs, LLaVA-1.5-7B~\cite{liu2023visual} and Qwen3-VL-8B~\cite{bai2025qwen3}, and collect per-head source-allocation vectors under two hallucination protocols. For the generative protocol, COCO-500, we sample 500 COCO~\cite{lin2014microsoft} val2014 images, generate captions with greedy decoding, and label generated COCO-object nouns as faithful or hallucinated by matching against the ground-truth object annotations in the CHAIR~\cite{rohrbach2018object} style. For the discriminative protocol, POPE-9000, we teacher-force both the ``Yes'' and ``No'' answer tokens on 9{,}000 yes/no questions~\cite{li2023evaluating} over the same 500 images. Findings~1--3 are evaluated on all four model--task combinations and reported as the mean over five fixed random seeds that control data splitting and probe training. We report the core conclusions in the three findings that follow, with the full experimental setup and complete results in the supplementary material.

\subsection{Finding 1: Attention Signals Are Head-Dependent}
\label{sec:finding1}

Prior work explains hallucination from various perspectives. For example, PAI~\cite{liu2024paying} emphasizes insufficient use of visual evidence, VCD~\cite{leng2024mitigating} attributes it to language-prior dominance, HGAI~\cite{jiang2025devils} identifies insufficient mid-layer visual attention on real objects, and OPERA~\cite{huang2024opera} focuses on anomalous dependence on specific anchor tokens. We take a per-head view of these phenomena and test whether the head-level signatures can serve as stable hallucination signals.

\begin{figure}[t]
    \centering
    \includegraphics[width=\linewidth]{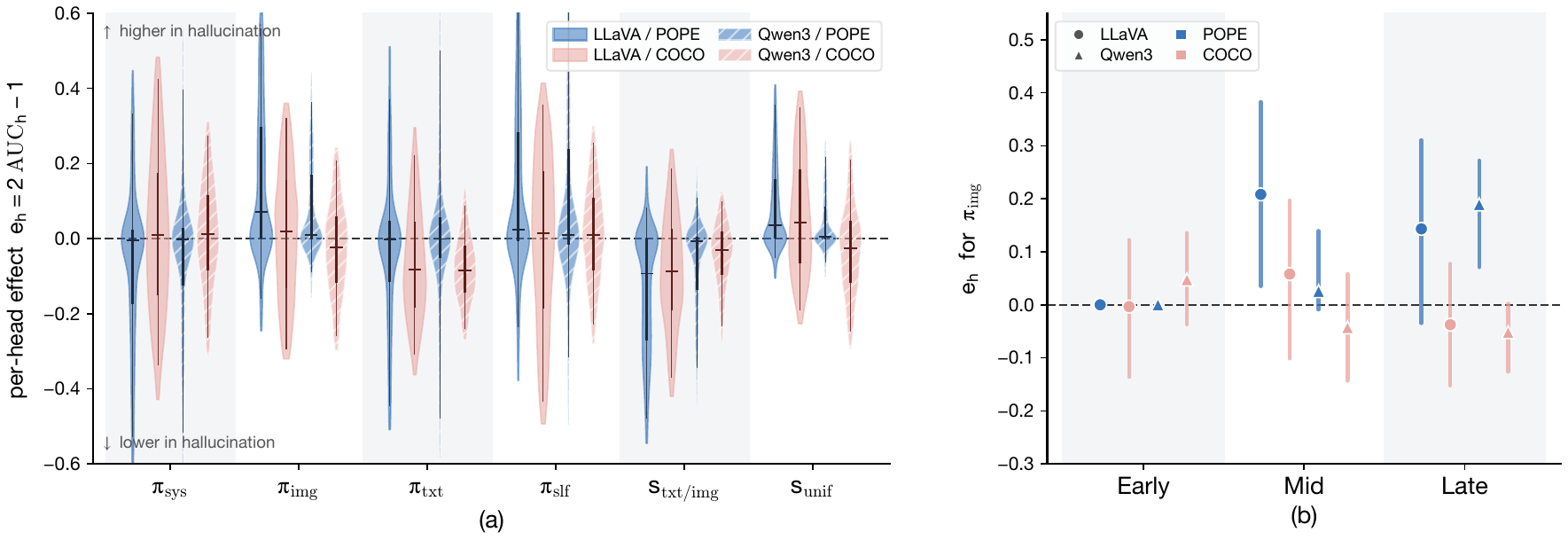}
    \caption{Per-head directional effect across six indicators (a) and across normalized-depth layer bands for $\pi_{\mathrm{img}}$ (b).}
    \label{fig:f1_view_dependent}
\end{figure}

\begin{figure}[t]
    \centering
    \includegraphics[width=\linewidth]{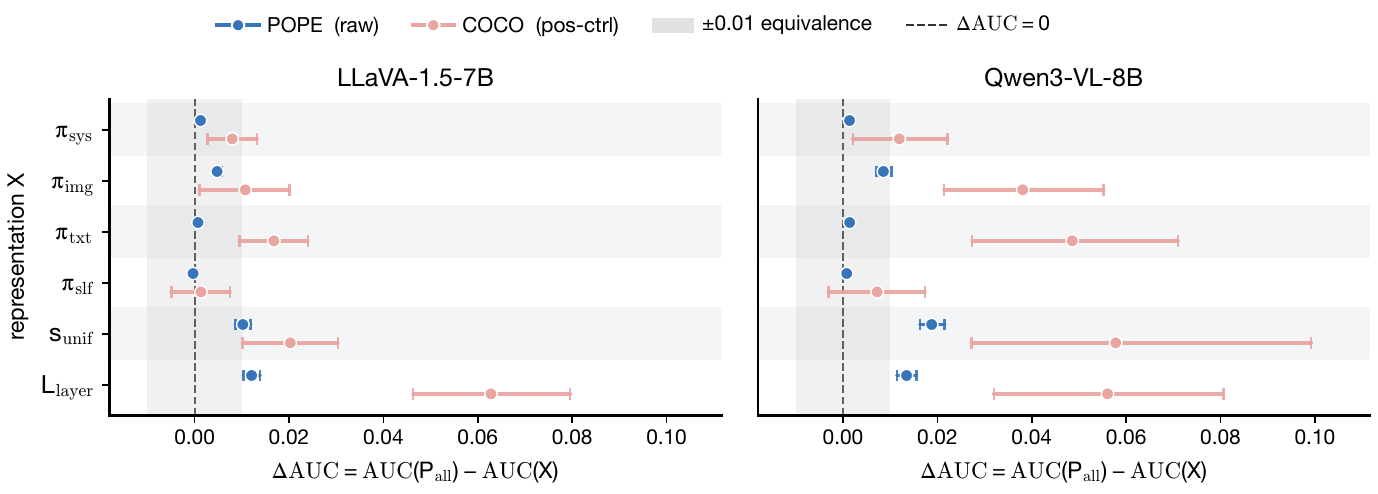}
    \caption{Predictive sufficiency of head-resolved representations. Gray band is the $\pm 0.01$ equivalence margin.}
    \label{fig:f1_ladder}
\end{figure}

We construct six representative head-level indicators derived from $\boldsymbol{\pi}_k$, including the four source channels $\pi_{\mathrm{sys}}, \pi_{\mathrm{img}}, \pi_{\mathrm{txt}}, \pi_{\mathrm{slf}}$, together with two composite features, text-to-image ratio and image-attention uniformity. We evaluate these indicators along two dimensions. For direction stability, we compute a signed AUC at each head independently, and stress-test the direction under generation-position and layer-range controls. For predictive sufficiency, we compare cross-validated AUC of single-feature head-resolved probes, a layer-aggregated pooling probe, and the multi-source head-resolved reference $P_{\mathrm{all}}$. The main results are shown in Figures~\ref{fig:f1_view_dependent} and~\ref{fig:f1_ladder}.

\begin{itemize}
    \item \textbf{The directional association of attention features is conditional.} As shown in Figure~\ref{fig:f1_view_dependent}(a), most directional-effect distributions straddle zero, meaning that within a single setting individual heads can carry opposite associations. Taking image attention as an example, its direction flips between POPE-9000 and COCO-500, and its peak-layer band shifts across models (Figure~\ref{fig:f1_view_dependent}(b)). No single feature therefore admits a direction that is uniformly stable across heads, models, tasks, and layer ranges.
    \item \textbf{Hallucination information lives in per-head organization.} As shown in Figure~\ref{fig:f1_ladder}, which single feature suffices is setting-dependent, but aggregating heads into a per-layer mean pooling consistently drops AUC across all settings, with the largest loss on COCO-500. Preserving head identity is therefore what carries the predictive information.
\end{itemize}

\noindent\textbf{Conclusion.} Overall, no single attention feature admits a uniformly stable direction, and layer-wise aggregation consistently loses information. This suggests that hallucination-related information is not carried by the global rise or fall of any single attention feature, but is primarily stored in the specific pattern of changes across individual attention heads.

\subsection{Finding 2: Hallucinations Break Stable Head Roles}
\label{sec:finding2}

Building on Finding~1, we further ask whether the per-head changes across different attention heads jointly reflect a stable, reproducible, and coherent internal phenomenon.

To answer this question, we first examine whether individual attention heads exhibit stable and mutually distinct behavior patterns under faithful generation, and then examine whether hallucination departs from these patterns in a reproducible way that concentrates on specific attention heads.

For each head $(l,h)$ we estimate the mean source-allocation over faithful tokens from the training split,
\begin{equation}
    \boldsymbol{\mu}^{F}_{l,h} \;=\; \mathbb{E}_{y_k \sim \mathcal{F}}\bigl[\boldsymbol{\pi}_{k,l,h}\bigr] \;\in\; \mathbb{R}^4,
    \label{eq:faithful_role}
\end{equation}
and define it as the head's faithful role. The corresponding standard deviation $\boldsymbol{\sigma}^{F}_{l,h}$ characterizes its natural fluctuation.

We validate this construction along two dimensions. For role stability, we compute the split-half Pearson correlation of $\boldsymbol{\mu}^{F}$ and the across-head heterogeneity-to-noise ratio. For deviation reproducibility, we position-match each hallucinated token to its nearest faithful counterpart and measure two cross-split statistics: the split-half cosine $T_{\cos}$ of the per-head studentized effect and the top-$10\%$ head energy share $C_{10}$, both tested against $1{,}000$-draw group-clustered sign-flip permutations. Results are as follows.

\begin{itemize}
    \item \textbf{Faithful roles are highly stable across all models and tasks.} The split-half correlation of $\boldsymbol{\mu}^{F}$ is $\geq 0.9999$ in every setting, and head-to-head differences are $98$--$853\times$ larger than the estimation noise, meaning individual heads have stable and distinct source-allocation baselines.
    \item \textbf{Hallucination-induced per-head deviations are highly consistent across independent splits.} $T_{\cos} = 0.757$--$0.996$ across the four (model, task) settings, and the top $10\%$ of heads selected on one split explain $21\%$--$33\%$ of the deviation energy on the other split, significantly higher than the $10\%$--$12\%$ null. After residualizing normal position trends from faithful generation, both statistics change by roughly $0.03$.
\end{itemize}

\begin{figure}[t]
    \centering
    \includegraphics[width=\linewidth]{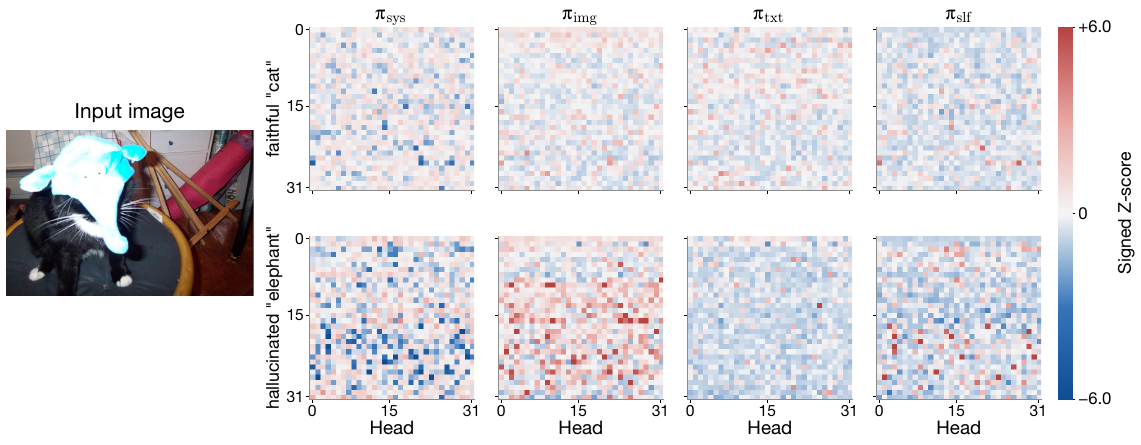}
    \caption{Per-head deviation from the faithful role for a faithful vs.\ hallucinated mention from the same caption.}
    \label{fig:f2_case}
\end{figure}

Figure~\ref{fig:f2_case} shows a concrete case. The faithful token ``cat'' stays close to $\boldsymbol{\mu}^{F}$ on all four channels, appearing as mild, directionless fluctuations across the $L \times H$ head grid, while the hallucinated token ``elephant'' shows clear departures whose magnitudes are large, directionally structured, and concentrated on a subset of heads.

\noindent\textbf{Conclusion.} These results show that under faithful generation individual attention heads carry stable and heterogeneous source-allocation roles, and that hallucination departs from these roles in a way that is reproducible across samples and concentrated on specific heads. We term this structured departure \textbf{Role-Break}.

\subsection{Finding 3: Role-Break Is Linearly Readable}
\label{sec:finding3}

Finding~2 established that hallucinations induce reproducible role departures concentrated on a subset of attention heads. We now examine whether these departures form a linearly readable hallucination signal, and in what structural form their predictive information is carried.

Concretely, for each head we normalize the current source-allocation against its faithful role:
\begin{equation}
    \rsrcH{l,h}(y_k) \;=\; \frac{\boldsymbol{\pi}_{k,l,h} - \boldsymbol{\mu}^{F}_{l,h}}{\boldsymbol{\sigma}^{F}_{l,h} + \varepsilon}.
    \label{eq:rsrc}
\end{equation}

We concatenate the coordinates across all heads and fit a single $L_2$-regularized logistic regression, evaluated under $5$-fold cross-validation. To localize where the predictive information resides, we further disrupt the coordinate identity of heads or source channels and re-evaluate after matching the overall deviation magnitude.

\begin{figure}[t]
    \centering
    \includegraphics[width=0.95\linewidth]{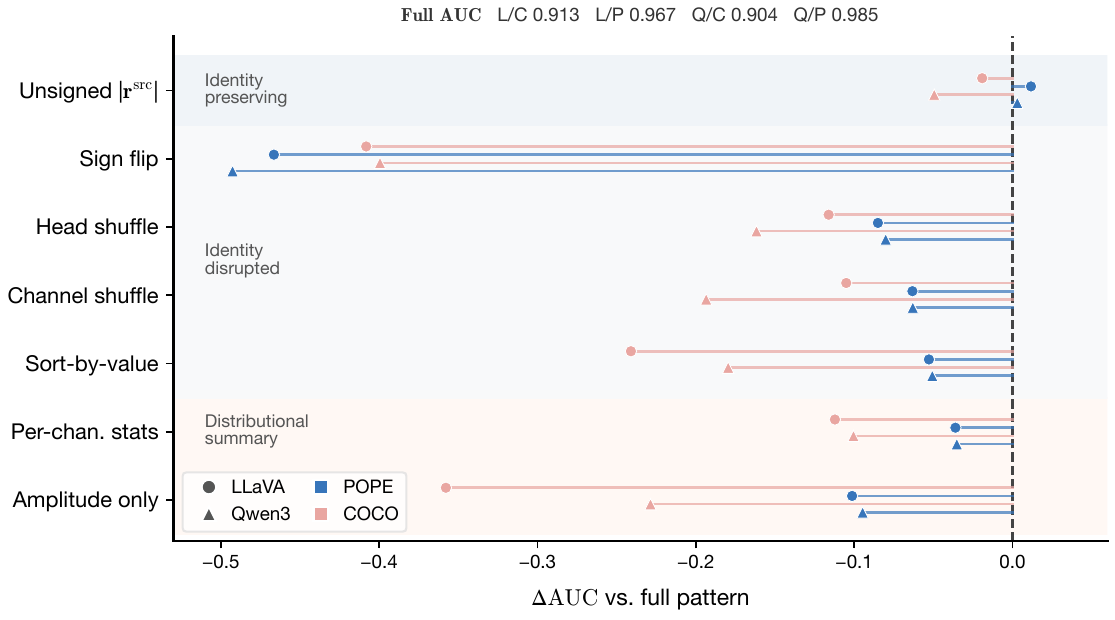}
    \caption{$\Delta$AUC of seven structural disruptions.}
    \label{fig:f2_anatomy}
\end{figure}

As shown in Figure~\ref{fig:f2_anatomy}, Role-Break is linearly readable on all four (model, task) settings, reaching AUC $0.904$ to $0.985$. Shuffling head or source-channel identity, or compressing the departures into aggregate summaries, all substantially reduce predictive performance. This indicates that Role-Break's information is carried by the per-head, per-source coordinate structure rather than by overall deviation magnitude. An amplitude-matched control in the supplementary material further confirms this conclusion.

\noindent\textbf{Conclusion.} Role-Break constitutes a linearly readable, per-head hallucination signature. Its discriminative power is carried by the deviation pattern across attention heads and source channels, and the faithful role provides an interpretable baseline for this predictive pattern.

\section{Role-Break-Based Hallucination Detection}
\label{sec:detection}

\begin{figure*}[t]
\centering
\includegraphics[width=\linewidth]{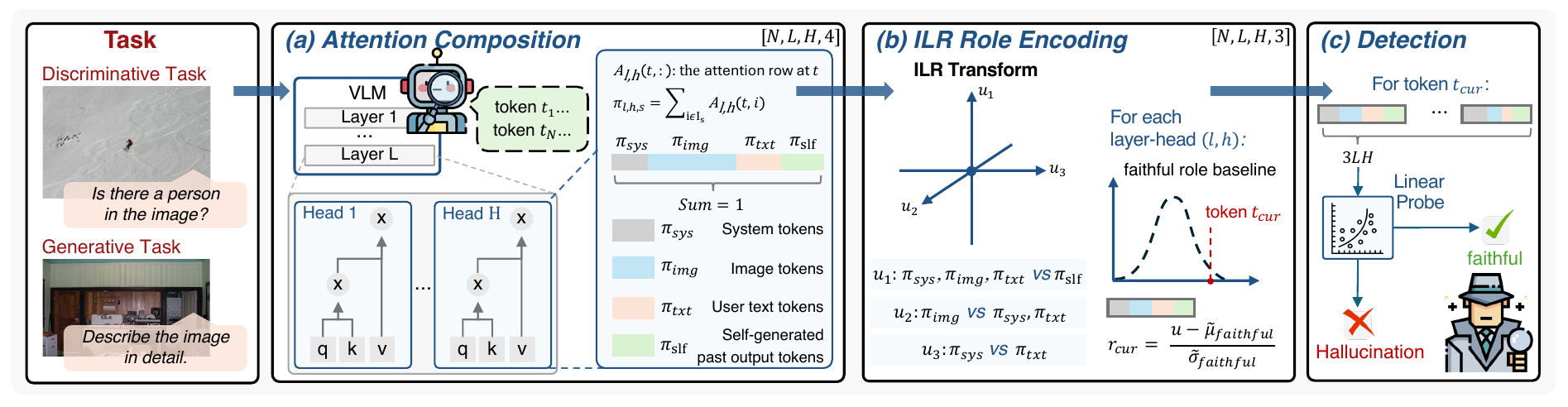}
\caption{Overview of the Role-Break-based hallucination detection framework.}
\label{fig:methoverview}
\end{figure*}

\subsection{Detector Design}

Building on the Role-Break phenomenon established above, we design a lightweight hallucination detector, whose overall pipeline is illustrated in Figure~\ref{fig:methoverview}. The method consists of three stages. (a) Attention Composition extracts the four-source attention composition of the current token from every layer-head. (b) ILR Role Encoding encodes each composition as a standardized deviation from the head-specific faithful role. (c) Detection jointly classifies the token from the deviation patterns of all heads.

Notably, the same pipeline applies to both discriminative and generative tasks, differing only in which token is scored. On discriminative tasks we score the ``Yes''/``No'' answer token. On generative tasks we score the content tokens of interest along the response. For each evaluated token $y_k$ at sequence position $t_k$, a single forward pass yields all attention statistics needed across the $L\times H$ layer-heads.

\paragraph{(a) Attention Composition.}
Using the per-head source-allocation vector $\boldsymbol{\pi}_{k,l,h}$ defined in Eq.~\ref{eq:vpartition}, each head is described by its four-source attention composition summing to one. Across $N$ evaluated tokens, the compositions form a tensor of size $N\times L\times H\times 4$. To handle structural and numerical zeros, we add $\varepsilon$ and re-normalize before the ILR transform.

\paragraph{(b) ILR Role Encoding.}
While Finding~3 works in the source-aligned $\boldsymbol{\pi}$ space to keep each coordinate semantically interpretable, for prediction we remove the sum-to-one redundancy on the simplex by applying the isometric log-ratio (ILR) transform~\citep{egozcue2003isometric} to map each four-part composition into three-dimensional Euclidean space. We adopt a hierarchical set of three balances that capture, respectively, context versus self-generated history, image versus textual context, and system prompt versus user text:
\begin{align}
u_1 &= \sqrt{\tfrac{3}{4}}\,\log\frac{(\pi_{\mathrm{sys}}\pi_{\mathrm{img}}\pi_{\mathrm{txt}})^{1/3}}{\pi_{\mathrm{slf}}}, \\
u_2 &= \sqrt{\tfrac{2}{3}}\,\log\frac{\pi_{\mathrm{img}}}{(\pi_{\mathrm{sys}}\pi_{\mathrm{txt}})^{1/2}}, \\
u_3 &= \sqrt{\tfrac{1}{2}}\,\log\frac{\pi_{\mathrm{sys}}}{\pi_{\mathrm{txt}}},
\end{align}
forming $\boldsymbol{u}_{k,l,h} \in \mathbb{R}^3$.
For each fixed layer-head position $(l,h)$, we estimate the empirical mean and standard deviation of its ILR coordinates on training-split faithful tokens, denoted $(\tilde{\boldsymbol{\mu}}^{F}_{l,h}, \tilde{\boldsymbol{\sigma}}^{F}_{l,h})$, and define the ILR-coordinate residual of the current token as:
\begin{equation}
\boldsymbol{r}_{k,l,h} = (\boldsymbol{u}_{k,l,h} - \tilde{\boldsymbol{\mu}}^{F}_{l,h}) \oslash \tilde{\boldsymbol{\sigma}}^{F}_{l,h}\in\mathbb{R}^3,
\end{equation}
where $\oslash$ denotes element-wise division. In ILR coordinates, the faithful role of a layer-head is characterized by this empirical mean and standard deviation, reflecting its typical relative balance across the four sources. Role-Break is the current token's directional deviation from that reference.

\paragraph{(c) Detection.}
Each evaluated token forms an independent sample. We flatten and concatenate its three-dimensional residuals across all layer-head positions into a single vector $\boldsymbol{r}_k = \operatorname{vec}_{l,h}(\boldsymbol{r}_{k,l,h}) \in\mathbb{R}^{3LH}$, and feed it into an $L_2$-regularized logistic-regression probe:
\begin{equation}
p_k = \sigma(\boldsymbol{w}^\top \boldsymbol{r}_k + b),
\end{equation}
where $p_k$ is the predicted probability that $y_k$ is hallucinated.

\begin{table*}[t]
    \small
    \centering
    \tabcolsep=1.0mm
    \begin{tabular}{llcccccccccccc}
    \toprule
    \multirow{2}{*}{\textbf{Benchmark}}
    & \multirow{2}{*}{\textbf{Method}}
    & \multicolumn{2}{c}{\textbf{MiniGPT-4}}
    & \multicolumn{2}{c}{\textbf{LLaVA-1.5}}
    & \multicolumn{2}{c}{\textbf{LLaVA-1.5}$^{\dagger}$}
    & \multicolumn{2}{c}{\textbf{InstructBLIP}}
    & \multicolumn{2}{c}{\textbf{Qwen3-VL}}
    & \multicolumn{2}{c}{\textbf{Qwen3.5}} \\
    \cmidrule(lr){3-4} \cmidrule(lr){5-6} \cmidrule(lr){7-8} \cmidrule(lr){9-10} \cmidrule(lr){11-12} \cmidrule(lr){13-14}
    &
    & \textbf{A-ROC} & \textbf{A-PR}
    & \textbf{A-ROC} & \textbf{A-PR}
    & \textbf{A-ROC} & \textbf{A-PR}
    & \textbf{A-ROC} & \textbf{A-PR}
    & \textbf{A-ROC} & \textbf{A-PR}
    & \textbf{A-ROC} & \textbf{A-PR} \\
    \midrule
    \multicolumn{14}{l}{\textit{Discriminative Benchmarks}} \\
    \midrule

    \multirow{7}{*}{POPE}
    & AvgEnt         & 51.24 & 51.34 & 80.14 & 84.19 & 68.80 & 75.58 & 70.82 & 67.92 & 76.70 & 78.45 & 82.10 & 86.01 \\
    & AvgProb        & 72.79 & 74.18 & 93.30 & 92.90 & 93.90 & 93.32 & 82.88 & 85.17 & 90.19 & 92.58 & 95.24 & 95.47 \\
    & RepProbing     & 87.41 & 86.95 & \underline{95.20} & \underline{95.35} & \underline{95.96} & \underline{95.96} & 92.09 & 92.25 & \underline{96.75} & \underline{96.66} & 96.66 & 96.65 \\
    & MetaToken      & 85.75 & 84.54 & 93.61 & 93.49 & 93.09 & 93.55 & 93.91 & 94.06 & 93.21 & 93.22 & \underline{96.86} & \underline{96.98} \\
    & DHCP           & 87.55 & 87.29 & 93.77 & 93.79 & 93.96 & 94.21 & 94.23 & \underline{94.26} & 85.42 & 75.08 & 94.01 & 94.23 \\
    & VIB-Probe      & \underline{92.85} & \underline{91.21} & 94.93 & 92.84 & 95.80 & 94.12 & \underline{95.02} & 93.47 & 95.91 & 94.04 & 96.16 & 94.80 \\
    & \textbf{Ours}  & \textbf{93.07} & \textbf{92.96} & \textbf{96.98} & \textbf{96.98} & \textbf{97.19} & \textbf{97.26} & \textbf{95.94} & \textbf{95.98} & \textbf{97.85} & \textbf{97.93} & \textbf{97.70} & \textbf{97.73} \\
    \midrule

    \multirow{7}{*}{AMBER}
    & AvgEnt         & 45.01 & 48.18 & 65.22 & 69.68 & 70.22 & 76.36 & 59.59 & 59.09 & 72.00 & 71.35 & 80.26 & 80.67 \\
    & AvgProb        & 52.09 & 53.83 & 90.48 & 90.66 & 92.66 & 92.58 & 64.63 & 66.67 & 87.86 & 89.21 & 95.08 & 95.17 \\
    & RepProbing     & 92.46 & \underline{91.86} & \underline{97.09} & \underline{97.20} & \underline{97.58} & \underline{97.70} & 95.24 & 95.37 & \underline{98.45} & \underline{98.51} & \underline{98.38} & \underline{98.31} \\
    & MetaToken      & 88.91 & 89.10 & 96.41 & 96.60 & 97.18 & 97.28 & 95.67 & \underline{95.55} & 97.57 & 97.68 & 97.63 & 97.78 \\
    & DHCP           & 91.50 & 91.83 & 93.89 & 94.08 & 95.40 & 95.61 & 94.46 & 94.64 & 93.54 & 93.73 & 94.43 & 94.15 \\
    & VIB-Probe      & \textbf{94.14} & 91.60 & 96.44 & 94.55 & 96.70 & 94.54 & \underline{95.80} & 93.44 & 97.08 & 94.35 & 97.79 & 96.17 \\
    & \textbf{Ours}  & \underline{94.04} & \textbf{94.32} & \textbf{98.11} & \textbf{98.20} & \textbf{98.55} & \textbf{98.59} & \textbf{96.43} & \textbf{96.16} & \textbf{98.92} & \textbf{98.97} & \textbf{98.59} & \textbf{98.67} \\
    \midrule
    \multicolumn{14}{l}{\textit{Generative Benchmarks}} \\
    \midrule

    \multirow{7}{*}{M-HalDetect}
    & AvgEnt         & 59.66 & 38.67 & 51.19 & 30.77 & 49.64 & 29.92 & 57.06 & 34.84 & 48.31 & 30.08 & 48.95 & 31.29 \\
    & AvgProb        & 47.81 & 29.47 & 50.93 & 31.39 & 49.18 & 30.10 & 50.53 & 28.60 & 45.85 & 26.91 & 50.88 & 33.89 \\
    & RepProbing     & 73.67 & 54.59 & 76.30 & 58.75 & 74.92 & 56.09 & 74.82 & 56.50 & 74.55 & 53.15 & 75.18 & 55.58 \\
    & MetaToken      & 79.85 & 66.13 & 81.47 & 68.68 & 80.38 & 66.85 & 80.90 & 68.62 & 81.03 & 64.06 & 78.75 & 60.64 \\
    & DHCP           & 78.97 & 65.83 & 72.80 & 52.88 & 72.43 & 52.56 & \underline{81.80} & \underline{69.07} & 72.44 & 49.65 & 69.42 & 44.65 \\
    & VIB-Probe      & \underline{81.82} & \underline{69.28} & \underline{82.57} & \underline{71.86} & \underline{83.74} & \underline{72.38} & 81.21 & 68.10 & \underline{84.86} & \underline{73.27} & \textbf{82.78} & \textbf{70.27} \\
    & \textbf{Ours}  & \textbf{83.64} & \textbf{73.19} & \textbf{85.19} & \textbf{75.74} & \textbf{85.83} & \textbf{76.40} & \textbf{83.72} & \textbf{72.43} & \textbf{85.93} & \textbf{75.33} & \underline{82.32} & \underline{68.47} \\
    \midrule

    \multirow{7}{*}{COCO-Caption}
    & AvgEnt         & 57.42 & 14.94 & 48.18 & 20.67 & 49.73 & 20.06 & 56.59 & 25.17 & 49.88 & 9.00 & 55.56 & 16.94 \\
    & AvgProb        & 57.67 & 15.03 & 48.56 & 21.30 & 50.16 & 20.70 & 56.40 & 25.77 & 46.69 & 7.25 & 54.43 & 16.37 \\
    & RepProbing     & 89.96 & 69.21 & 88.39 & 69.52 & 87.55 & 68.37 & 90.13 & 77.45 & 90.50 & 74.72 & \underline{97.27} & \underline{92.26} \\
    & MetaToken      & 87.21 & 60.77 & 88.44 & 69.84 & 88.00 & 66.96 & 90.41 & 75.77 & 92.87 & 73.97 & 92.00 & 70.68 \\
    & DHCP           & \underline{91.35} & 65.43 & 81.95 & 58.90 & 79.05 & 53.69 & \underline{93.34} & 81.86 & 65.00 & 27.28 & 66.28 & 28.60 \\
    & VIB-Probe      & 90.88 & \underline{71.24} & \underline{91.01} & \underline{77.81} & \underline{90.74} & \underline{76.12} & 92.42 & \underline{83.03} & \underline{96.10} & \underline{79.23} & \textbf{97.37} & \textbf{93.47} \\
    & \textbf{Ours}  & \textbf{93.97} & \textbf{73.72} & \textbf{92.93} & \textbf{80.53} & \textbf{93.89} & \textbf{81.79} & \textbf{93.74} & \textbf{83.34} & \textbf{96.97} & \textbf{82.93} & 96.04 & 85.40 \\

    \bottomrule
    \multicolumn{14}{l}{\footnotesize $\dagger$ 13B-scale variant.}
    \end{tabular}
    \caption{Hallucination detection performance (AUROC / AUPRC). Values are means over three seeds.}
    \label{tab:detection_main}
\end{table*}

\subsection{Experimental Setup}
\label{sec:setup}

Following prior work~\cite{zhang2026vib, li2024reference}, we evaluate the effectiveness of our detector under a unified suite of benchmarks, metrics, and evaluation protocol.

\paragraph{Benchmarks.}
We evaluate our detector on four hallucination benchmarks. \textbf{POPE}~\cite{li2023evaluating} probes object-level existence with yes/no questions in three subsets (\emph{Random}, \emph{Popular}, \emph{Adversarial}), each containing $3{,}000$ questions on $500$ MS COCO~\cite{lin2014microsoft} val2014 images. \textbf{AMBER}~\cite{wang2023amber} extends the yes/no protocol to attribute and relation queries, and we use all $14{,}216$ discriminative queries over $1{,}004$ images. \textbf{M-HalDetect}~\cite{gunjal2024detecting} provides span-level expert annotations on VLM-generated captions across $786$ images. \textbf{COCO-Caption} additionally evaluates free-form captions on about $2{,}000$ sampled COCO~\cite{lin2014microsoft} val2014 images, where generated COCO-object nouns are labeled by matching against the ground-truth object annotations in the CHAIR~\cite{rohrbach2018object} style.

\paragraph{Base Models.}
We evaluate our detector on six representative VLMs selected along two dimensions. \emph{(i) Visual-fusion architecture:} MiniGPT-4-7B~\cite{zhu2024minigpt}, LLaVA-1.5-7B~\cite{liu2023visual}, InstructBLIP-Vicuna-7B~\cite{dai2023instructblip}, Qwen3-VL-8B~\cite{bai2025qwen3}, and the latest Qwen3.5-9B. \emph{(ii) Model scale:} we additionally include LLaVA-1.5-13B to cover a larger backbone.

\paragraph{Baselines.}
We compare our method with four representative families of hallucination detectors: uncertainty-based scores including AvgProb and AvgEnt, representation probes on hidden states or per-head output vectors including RepProbing and VIB-Probe~\cite{zhang2026vib}, the attention-based DHCP~\cite{zhang2025dhcp}, and the token-statistics-based MetaToken~\cite{fieback2024metatoken}. Table~\ref{tab:detector_compare} summarizes these baselines along signal source, feature dimensionality, probe parameter count, and probe type.

\paragraph{Metrics.}
For detection, we report the area under the ROC curve (AUROC) and the area under the precision--recall curve (AUPRC) on all benchmarks. For mitigation, we additionally report Macro-F1 on the discriminative benchmarks.

\paragraph{Implementation Details.}
Unless otherwise stated, all experiments in this section (including main results, ablations, and mitigation) uniformly use three fixed random seeds $\{42, 43, 44\}$ under $80/20$ image-disjoint splits, and we report the mean AUROC / AUPRC across seeds. Detailed implementation setup, per-seed standard deviations, and complete results are provided in the supplementary material.

\begin{table}[t]
    \small
    \centering
    \setlength{\tabcolsep}{2.5pt}
    \begin{tabular}{llrrl}
    \toprule
    \textbf{Method} & \textbf{Signal} & \textbf{Feat.} & \textbf{Para.} & \textbf{Probe} \\
    \midrule
    AvgProb        & forced-token logprob        & 1              & 0              & N/A            \\
    AvgEnt         & token entropy               & 1              & 0              & N/A            \\
    RepProbing     & last hidden state           & 4{,}096        & 4{,}097        & Linear         \\
    MetaToken      & scalars + head attn         & 1{,}034        & 1{,}035        & Linear         \\
    DHCP           & cross-modal attn map        & 590K           & 75.5M          & MLP            \\
    VIB-Probe      & per-head output vec         & 131K           & 135.3M         & MLP+VIB        \\
    \textbf{Ours}  & \textbf{source alloc.\ ILR} & \textbf{3{,}072} & \textbf{3{,}073} & \textbf{Linear} \\
    \bottomrule
    \end{tabular}
    \caption{Comparison of baselines on LLaVA-1.5-7B.}
    \label{tab:detector_compare}
\end{table}

\begin{figure*}[!t]
\centering
\includegraphics[width=\linewidth]{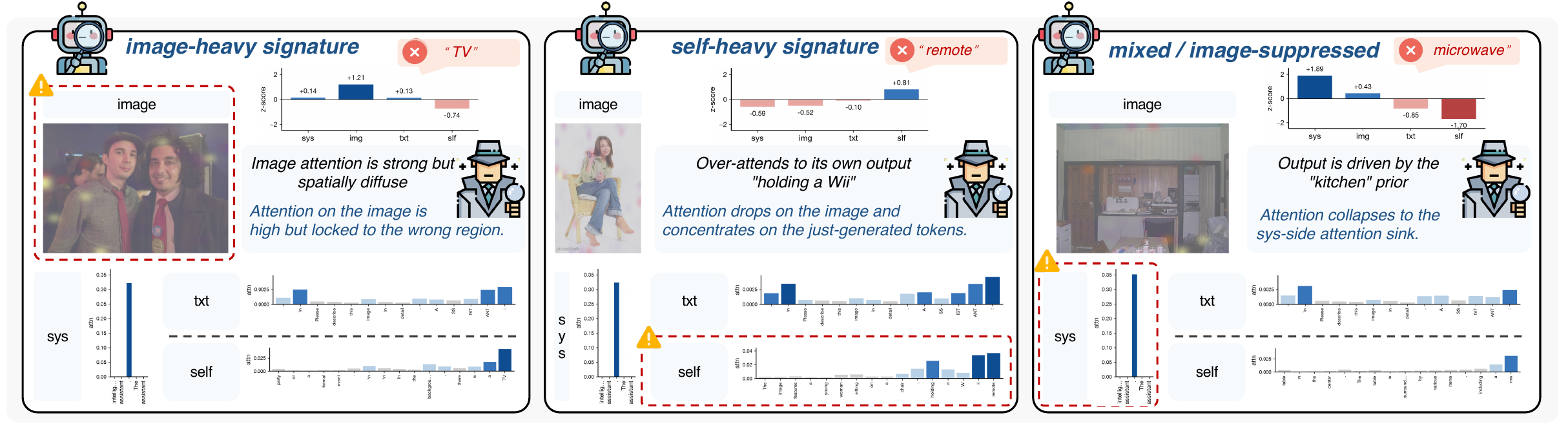}
\caption{Case study: one representative hallucinated token per Role-Break cluster (K-means, $K\!=\!3$) on LLaVA-1.5-7B.}
\label{fig:case_study}
\end{figure*}

\subsection{Main Detection Results}

Table~\ref{tab:detection_main} reports the detection performance across six VLMs and four benchmarks. Our method achieves the best results on 21 out of the 24 (VLM, benchmark) combinations, and remains highly competitive on the remaining 3.

\begin{itemize}
    \item \textbf{Task-type analysis.} Baselines behave very differently across the two task types. RepProbing is strong on the discriminative benchmarks but degrades sharply on the generative ones, with its AUPRC dropping from around $95.2$ to $65.5$ on average. VIB-Probe and DHCP remain strong on both task types. Nevertheless, our method still delivers the best overall performance. Averaged over six VLMs, we reach $96.95$ AUROC and $96.98$ AUPRC on the discriminative benchmarks and $89.52$ AUROC, $77.44$ AUPRC on the generative ones, exceeding the strongest baseline VIB-Probe by more than $1$ point on AUROC and $2$--$3$ points on AUPRC in both regimes.

    \item \textbf{Architectures and scales analysis.} Across visual-fusion architectures, our method leads on most of them. In particular, Qwen3.5-9B is a hybrid-attention model with $32$ decoder layers of which only $8$ are full-attention, so we retain source-allocation vectors only from these $8$ layers with $16$ attention heads each, giving $128$ layer-head positions and a $384$-dimensional feature space. VIB-Probe still leads marginally on a few generative metrics thanks to its high-dimensional hidden states, but our method stays within about $1.5$ AUROC points while using an $85\times$ smaller feature and $90{,}000\times$ fewer parameters. Across model scales, scaling from LLaVA-1.5-7B to the $13$B variant yields consistent gains, indicating that head-level allocation signals become clearer in larger models.
\end{itemize}

Notably, our setup follows prior work~\cite{zhang2026vib,li2024reference} in teacher-forcing part of the answer tokens. In the supplementary material we report within-polarity AUROC and an answer-token-identity baseline to rule out any confound from this coupling, together with a sample-efficiency analysis showing that a few hundred labeled faithful tokens per class recover most of the full-training performance.

\begin{table}[!t]
\centering
\small
\setlength{\tabcolsep}{3pt}
\begin{tabular}{llcccc}
\toprule
\textbf{Axis} & \textbf{Variant} & \textbf{POPE} & \textbf{AMBER} & \textbf{M-Hal} & \textbf{COCO} \\
\midrule
Full & --      & \textbf{96.5} & \textbf{97.4} & \textbf{84.4} & \textbf{94.6} \\
\midrule
\multirow{4}{*}{Channels}
 & $\pi_{\mathrm{sys}}$ only         & 88.3 & 89.0 & 77.4 & 85.6 \\
 & $\pi_{\mathrm{img}}$ only         & 95.4 & 95.7 & 82.2 & 93.2 \\
 & $\pi_{\mathrm{txt}}$ only         & 96.1 & 96.7 & 81.1 & 91.7 \\
 & $\pi_{\mathrm{slf}}$ only         & 88.5 & 89.3 & 82.7 & 91.2 \\
\midrule
\multirow{4}{*}{Layers}
 & Shallow-25\%                    & 79.1 & 85.5 & 79.8 & 91.9 \\
 & Front-50\%                      & 95.6 & 96.3 & 83.3 & 93.9 \\
 & Mid-50\%                        & 96.2 & 97.1 & 83.6 & 93.5 \\
 & Deep-25\%                       & 95.5 & 95.7 & 80.7 & 90.2 \\
\midrule
\multirow{2}{*}{Probe}
 & MLP (1-layer)                   & 96.4 & 98.3 & 81.9 & 93.6 \\
 & MLP (2-layer)                   & 96.4 & 98.3 & 81.7 & 93.7 \\
\bottomrule
\end{tabular}
\caption{Ablation diagnostics: macro-average ROC-AUC (\%) across six VLMs and four benchmarks.}
\label{tab:ablation_minimal}
\end{table}

\subsection{Ablation Studies}
\label{sec:ablation}

We conduct ablations across three dimensions and report the macro-average AUC across all six VLMs and four benchmarks. The Full configuration is the setting used in the main table. As summarized in Table~\ref{tab:ablation_minimal}, the Full configuration performs best across all four benchmarks. Concretely, in terms of source channels, no single channel matches the four-channel probe. $\pi_{\mathrm{img}}$ and $\pi_{\mathrm{txt}}$ are the strongest solo channels while $\pi_{\mathrm{sys}}$ and $\pi_{\mathrm{slf}}$ are weaker in isolation, and the full probe outperforms every solo channel by up to $9$ points, confirming that Role-Break spans multiple source channels. In terms of layer coverage, any half of the network already recovers most of the signal, while full-depth aggregation is optimal. In terms of probe capacity, linear probes match or outperform MLPs on POPE, M-HalDetect, and COCO, consistent with Role-Break being linearly readable.

\begin{table}[!t]
\centering
\small
\begin{tabular}{lcccc}
\toprule
\textbf{Method} & \textbf{LLaVA} & \textbf{LLaVA}$^{\dagger}$ & \textbf{Qwen3} & \textbf{Qwen3.5} \\
\midrule
P / Van.  & 86.35 & 87.44 & 89.53 & 89.68 \\
P / Ours  & 89.33 & 90.27 & 92.37 & 92.13 \\
\midrule
A / Van. & 82.29 & 83.07 & 87.91 & 87.60 \\
A / Ours & 90.17 & 92.44 & 93.94 & 92.35 \\
\bottomrule
\multicolumn{5}{l}{\footnotesize $\dagger$ 13B-scale variant.}
\end{tabular}
\caption{Detector-guided answer flipping on POPE (P, avg.\ of three subsets) and AMBER (A). Macro-F1 (\%).}
\label{tab:mitigation}
\end{table}

\subsection{Role-Break-Guided Mitigation}
\label{sec:mitigation}

We further verify whether the Role-Break signature is actionable at inference time. Following the mitigation setup of DHCP~\cite{zhang2025dhcp}, on the discriminative Yes/No protocols we flip the model's answer whenever the probe classifies it as hallucinated, with no threshold tuning. Table~\ref{tab:mitigation} reports macro-F1 on POPE and AMBER. Detector-guided flipping delivers consistent gains across all four VLMs, with an average improvement of $2.8$ F1 on POPE and $7.0$ F1 on AMBER. These results show that the detected tokens can be directly flipped in the discriminative setting, and that this simple use of the signal transfers across architectures.

Note that answer flipping applies only to the discriminative regime. On generative benchmarks, the space of alternative continuations is unconstrained, so we do not adopt the simple binary flipping as the mitigation strategy, and we leave a generative-side intervention to future work.

\subsection{Case Study}
\label{sec:case_study}

To illustrate Role-Break concretely, we cluster all hallucinated tokens produced by LLaVA-1.5-7B on the COCO-Caption benchmark with K-means ($K\!=\!3$, selected by silhouette) over their token-level source-allocation deviations from a position-matched faithful baseline, and visualize one representative token from each cluster in Figure~\ref{fig:case_study}. The three clusters exhibit qualitatively distinct source-allocation deviation patterns, including image-heavy, self-heavy, and image-suppressed. Thus Role-Break automatically captures multiple structurally distinct hallucination patterns under a head-level view, matching the mixed nature of hallucinations.

\section{Conclusion}
\label{sec:conclusion}

In this work, we propose faithful role and Role-Break as a head-level characterization of VLM hallucination. Through analysis, we reveal the necessity, the internal structure, and the appropriate representation of Role-Break. Building on this, we develop a simple linear hallucination detector on top of Role-Break that requires no additional training of the VLM and introduces no extra VLM inference steps, with a feature dimension below $5{,}000$ on all six VLMs, reaching an average AUROC of $93.23$ across six VLMs and four benchmarks. A small-scale intervention experiment further shows that the detected tokens can be directly acted upon in the discriminative setting.

Our approach is grounded in prior interpretability findings that trained VLMs develop functionally specialized attention heads~\cite{olsson2022context,wang2022interpretability}, which offers a plausible reason why per-head deviations from stable source-allocation baselines carry a discriminative signal. However, our current evidence cannot distinguish whether Role-Break is a cause of hallucination, a downstream response to it, or a partial self-correction attempt by the model, as we have not performed controlled head-level interventions. Establishing this mechanistic account, and extending Role-Break to more targeted mitigation strategies, are promising directions for future work.

\bibliography{refs}

\clearpage
\appendix

\twocolumn[
\begin{center}
{\Large\textbf{Supplementary Material}}
\vspace{1em}
\end{center}
]

\noindent This supplementary material provides (i) the complete numerical values and detailed experimental setups for the main-text findings, and (ii) additional experiments that validate the detector's design choices, rule out potential confounds, and confirm statistical significance.

\vspace{0.5em}
\noindent\textbf{Part I. Understanding Role-Break}
\begin{itemize}
    \item \textbf{\ref{app:role_validity}.} Numerical results for Findings 2 and 3, including role validity, anatomy, and amplitude-matched control.
    \item \textbf{\ref{app:ilr_robustness}.} Robustness verification showing that both structural claims replicate under the ILR representation.
\end{itemize}

\vspace{0.3em}
\noindent\textbf{Part II. Role-Break-Based Detection}
\begin{itemize}
    \item \textbf{\ref{app:detection_full}.} Complete detection and mitigation results with per-seed standard deviations and statistical significance tests.
    \item \textbf{\ref{app:signal_source}.} Signal source analysis comprising the ingredient ladder and the channel-choice sweep.
    \item \textbf{\ref{app:confound}.} Confound controls, including answer-polarity control and sample efficiency.
    \item \textbf{\ref{app:setup_full}.} Baseline implementations.
\end{itemize}

\vspace{0.3em}
\noindent\textbf{Conclusions.}
Beyond completeness, the additional experiments demonstrate four properties of the detector.

(i) \textbf{The ILR representation preserves performance while removing redundancy.} \ref{app:ilr_robustness} shows that both structural claims of Finding~3 replicate under the source-aligned and ILR representations, confirming that the ILR coordinate change removes the sum-to-one constraint without sacrificing predictive information.

(ii) \textbf{Head identity and the four-source partition are jointly necessary.} \ref{app:signal_source} decomposes the detector from two directions. The ingredient ladder shows that preserving per-head identity is the sole load-bearing step in the pipeline. The channel-choice sweep further shows that, holding the pipeline fixed, the four-source partition is the only per-head channel that makes hallucination linearly readable. Every alternative, including image attention, attention peakiness, attention concentration, value norm, and even the full head output vector at over $40\times$ the dimensionality, loses to it.

(iii) \textbf{The signal is not an artifact of the evaluation protocol.} \ref{app:confound} addresses two potential confounds. The answer-polarity control shows that within-polarity AUROC matches the full-POPE reference while a probe using only the forced-answer identity is exactly at chance, ruling out teacher-forcing leakage. The sample-efficiency analysis shows that a few hundred labeled faithful tokens per class suffice to recover most of the reported performance.

(iv) \textbf{The improvements are statistically significant.} \ref{app:detection_full} reports a one-sided Wilcoxon signed-rank test showing our method outperforms every baseline with $p < 10^{-9}$.

\section{Numerical Results for Findings}
\label{app:role_validity}

\subsection*{Finding 2: Role Validity}

This subsection reports the numerical evidence behind Finding~2: on each of the four (model, task) settings, the faithful role $(\boldsymbol{\mu}^{F}, \boldsymbol{\sigma}^{F})$ is highly reproducible across independent splits of the faithful pool, $\boldsymbol{\mu}^{F}$ is strongly heterogeneous across heads, and hallucination-induced deviations are reproducible across samples.

\paragraph{Split-half stability and across-head heterogeneity.} For each setting we randomly split the faithful pool into two halves, compute $(\boldsymbol{\mu}^{F}, \boldsymbol{\sigma}^{F})$ on each half, and report the Pearson correlation between them (all numbers are means over $5$ random splits with standard deviations $\le 3 \times 10^{-4}$). To quantify across-head heterogeneity, for each source channel $s$ we compute the ratio $\mathrm{SNR}(s) = \sigma_{\mathrm{across}}(s) / \sigma_{\mathrm{within}}(s)$, where $\sigma_{\mathrm{across}}$ is the across-head variation of $\boldsymbol{\mu}^{F}$ and $\sigma_{\mathrm{within}}$ is the mean per-head estimation noise. Table~\ref{tab:role_validity} reports both.

\begin{table}[h]
    \centering
    \small
    \setlength{\tabcolsep}{4pt}
    \begin{tabular}{lcccccc}
        \toprule
        & \multicolumn{2}{c}{Split-half $r$} & \multicolumn{4}{c}{SNR} \\
        \cmidrule(lr){2-3} \cmidrule(lr){4-7}
        & $\boldsymbol{\mu}^{F}$ & $\boldsymbol{\sigma}^{F}$ & $\pi_{\mathrm{sys}}$ & $\pi_{\mathrm{img}}$ & $\pi_{\mathrm{txt}}$ & $\pi_{\mathrm{slf}}$ \\
        \midrule
        L / C & $1.0000$ & $0.9991$ & $190\times$ & $218\times$ & $231\times$ & $137\times$ \\
        L / P & $1.0000$ & $0.9994$ & $579\times$ & $853\times$ & $520\times$ & $743\times$ \\
        Q / C & $0.9999$ & $0.9987$ & $122\times$ & $152\times$ & $98\times$  & $108\times$ \\
        Q / P & $1.0000$ & $0.9991$ & $639\times$ & $265\times$ & $616\times$ & $744\times$ \\
        \bottomrule
    \end{tabular}
    \caption{Role validity: split-half Pearson correlation of the faithful role (left) and across-head heterogeneity SNR per source channel (right). L~=~LLaVA-1.5-7B, Q~=~Qwen3-VL-8B, P~=~POPE-9000, C~=~COCO-500.}
    \label{tab:role_validity}
\end{table}

\paragraph{Deviation reproducibility.} To test whether hallucination-induced deviations form a reproducible spatial pattern across heads, we construct matched hallucinated--faithful pairs via one-to-one nearest-position matching within each caption (Hungarian algorithm, no faithful token reused). For each pair we compute the raw difference in source-allocation across all heads, then studentize using caption-clustered standard errors to obtain a per-head effect vector $\bt \in \mathbb{R}^{4 \times L \times H}$. We measure two statistics on independent split-halves of the caption groups: (i) $T_{\cos}$, the cosine similarity between the per-head studentized effects computed on each half, which measures whether the deviation pattern points in the same direction; and (ii) $C_{10}$, the fraction of total $\|\bt\|^2$ energy captured by the top-$10\%$ heads selected on one half and evaluated on the other, which measures whether deviations concentrate on specific heads. Both statistics are tested against a $1{,}000$-draw caption-clustered sign-flip null, where all pairs from the same caption receive the same random sign to preserve within-caption correlations.

\begin{figure}[t]
    \centering
    \includegraphics[width=\linewidth]{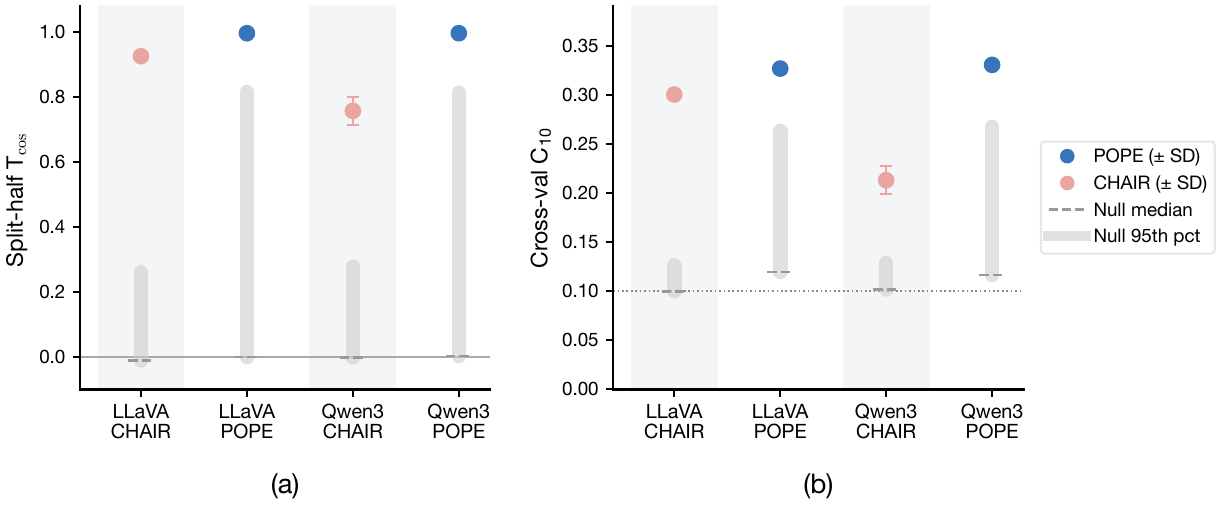}
    \caption{Split-half pattern reproducibility $T_{\cos}$ (a) and cross-validated top-$10\%$ head energy share $C_{10}$ (b), with caption-clustered sign-flip null bands.}
    \label{fig:f2_structured}
\end{figure}

Figure~\ref{fig:f2_structured} reports the results. $T_{\cos}$ ranges from $0.757$ to $0.996$, and $C_{10}$ ranges from $21\%$ to $33\%$, both well above their sign-flip null bands on every setting.

\subsection*{Finding 3: Anatomy and Controls}
\label{app:finding3}

The full-pattern AUC (5-seed out-of-fold means) reaches $0.913$ (L/C), $0.967$ (L/P), $0.904$ (Q/C), and $0.985$ (Q/P), confirming linear readability across both architectures and task formats.

\paragraph{Anatomy numerical values.}
Table~\ref{tab:f2_anatomy} reports the raw detection AUCs behind the anatomy plot of Finding~3 (Figure~4 in the main paper). Feature families are: F0 = $\rsrc$ (full signed pattern); F1 = $|\rsrc|$ (unsigned magnitude); F2 = sample sign flip; F3 = sample head shuffle; F4 = sample channel shuffle; F5 = sort-by-value; F6 = per-channel summary ($32$-dim); F7 = amplitude only ($1$-dim). All numbers are $5$-seed means with standard deviations $\le 0.03$.

\begin{table}[h]
    \centering
    \small
    \setlength{\tabcolsep}{5pt}
    \begin{tabular}{lcccc}
        \toprule
        & L / C & L / P & Q / C & Q / P \\
        \midrule
        F0 & $0.913$ & $0.967$ & $0.904$ & $0.985$ \\
        F1 & $0.893$ & $0.979$ & $0.854$ & $0.988$ \\
        F2 & $0.504$ & $0.501$ & $0.504$ & $0.492$ \\
        F3 & $0.796$ & $0.882$ & $0.742$ & $0.905$ \\
        F4 & $0.808$ & $0.904$ & $0.710$ & $0.922$ \\
        F5 & $0.672$ & $0.915$ & $0.724$ & $0.934$ \\
        F6 & $0.800$ & $0.931$ & $0.803$ & $0.950$ \\
        F7 & $0.555$ & $0.866$ & $0.675$ & $0.890$ \\
        \bottomrule
    \end{tabular}
    \caption{Detection AUC under seven feature families derived from $\rsrc(y_k)$ on the four (model, task) combinations.}
    \label{tab:f2_anatomy}
\end{table}

\paragraph{Amplitude-matched control.}
To verify that Role-Break's predictive power does not simply reflect ``larger deviations on hallucinated mentions,'' we match hallucinated and faithful instances on log-amplitude (bin edges are computed on the training fold only) and re-evaluate both the amplitude-only probe and the full-pattern probe on the matched pool. Before matching, the KS statistic ranges from $0.15$ to $0.65$ and the standardized mean difference (SMD) from $+0.13$ to $+1.75$. Table~\ref{tab:f3_amp_matched} reports the post-matching values with subscripts showing the reduction. After matching, the amplitude probe collapses to chance ($0.499$--$0.502$) on every setting, while the pattern probe retains AUC $0.876$--$0.942$. The predictive information therefore resides in how deviation magnitudes are distributed across specific heads and source channels, not in overall deviation size.

\begin{table}[h]
    \centering
    \small
    \setlength{\tabcolsep}{4pt}
    \begin{tabular}{lcccc}
        \toprule
        Setting & KS & SMD & Amp & Pattern \\
        \midrule
        L / C & $0.05_{-0.10}$ & $+0.02_{-0.11}$ & $0.499$ & $0.898$ \\
        L / P & $0.02_{-0.58}$ & $+0.01_{-1.56}$ & $0.501$ & $0.916$ \\
        Q / C & $0.08_{-0.21}$ & $-0.01_{-0.56}$ & $0.502$ & $0.876$ \\
        Q / P & $0.02_{-0.63}$ & $+0.01_{-1.74}$ & $0.502$ & $0.942$ \\
        \bottomrule
    \end{tabular}
    \caption{Amplitude-matched evaluation. KS and SMD columns are post-matching values with subscripts showing the reduction from pre-matching. Amp = amplitude-only probe AUC, Pattern = full-pattern probe AUC.}
    \label{tab:f3_amp_matched}
\end{table}

\section{ILR Robustness}
\label{app:ilr_robustness}

The main text uses the source-aligned representation $\rsrc \in \mathbb{R}^{4LH}$ in Finding~3 because its four coordinates retain the semantics of the four contextual sources. The detector (Section~4) instead uses $\rilr \in \mathbb{R}^{3LH}$, which removes the sum-to-one redundancy on the compositional simplex. To verify that Finding~3's conclusions do not depend on this choice, we re-run two key analyses under $\rilr$: (i) the anatomy of feature families and (ii) the amplitude-matched pattern AUC. The ILR baseline statistics $\boldsymbol{\mu}^{\mathrm{ilr}}_F, \boldsymbol{\sigma}^{\mathrm{ilr}}_F$ are estimated on training-fold faithful tokens only, matching the main-text protocol.

\paragraph{Anatomy.} Table~\ref{tab:app_anatomy_ilr} reports the same F0--F7 feature families as Table~\ref{tab:f2_anatomy}, recomputed under $\rilr$. The qualitative pattern is identical to the $\rsrc$ version, confirming that Finding~3's structural conclusion holds regardless of which representation is used.

\begin{table}[h]
    \centering
    \small
    \setlength{\tabcolsep}{5pt}
    \begin{tabular}{lcccc}
        \toprule
        & L / C & L / P & Q / C & Q / P \\
        \midrule
        F0 & $0.921$ & $0.951$ & $0.885$ & $0.967$ \\
        F1 & $0.900$ & $0.959$ & $0.831$ & $0.969$ \\
        F2 & $0.501$ & $0.508$ & $0.503$ & $0.514$ \\
        F3 & $0.567$ & $0.839$ & $0.550$ & $0.828$ \\
        F4 & $0.873$ & $0.941$ & $0.729$ & $0.951$ \\
        F5 & $0.605$ & $0.884$ & $0.535$ & $0.875$ \\
        F6 & $0.681$ & $0.911$ & $0.537$ & $0.939$ \\
        F7 & $0.460$ & $0.820$ & $0.310$ & $0.799$ \\
        \bottomrule
    \end{tabular}
    \caption{Anatomy under $\rilr$, same F0--F7 protocol as Table~\ref{tab:f2_anatomy}.}
    \label{tab:app_anatomy_ilr}
\end{table}

\paragraph{Amplitude matching.} Matching hallucinated and faithful tokens on log-amplitude of $\rilr$ (20 quantile bins, edges fit on the training fold) reduces the standardized mean difference to below $0.021$ on all four combinations. The amplitude-only probe collapses to $0.499$--$0.503$ on matched pairs, while the full pattern retains $0.811$--$0.943$ AUC (versus $0.885$--$0.967$ un-matched). The pattern beyond amplitude claim of Finding~3 therefore replicates under $\rilr$.

\paragraph{Summary.} Both structural claims replicate under $\rilr$ with matching directions and comparable effect sizes. The two representations anchor the same underlying Role-Break; the choice between them is one of geometric non-redundancy versus source semantics, not of what the analyses can conclude.

\begin{table}[h]
\centering
\small
\setlength{\tabcolsep}{4pt}
\begin{tabular}{llcc}
\toprule
\textbf{Benchmark} & \textbf{Model} & \textbf{AUROC} & \textbf{AUPRC} \\
\midrule
\multirow{6}{*}{POPE-Random}
 & MiniGPT-4      & $95.96_{\pm 0.31}$ & $95.87_{\pm 0.46}$ \\
 & LLaVA-1.5-7B   & $98.16_{\pm 0.15}$ & $98.22_{\pm 0.12}$ \\
 & LLaVA-1.5-13B  & $98.23_{\pm 0.20}$ & $98.26_{\pm 0.21}$ \\
 & InstructBLIP   & $97.62_{\pm 0.42}$ & $97.59_{\pm 0.47}$ \\
 & Qwen3-VL       & $98.86_{\pm 0.29}$ & $98.85_{\pm 0.36}$ \\
 & Qwen3.5        & $98.71_{\pm 0.31}$ & $98.73_{\pm 0.32}$ \\
\midrule
\multirow{6}{*}{POPE-Popular}
 & MiniGPT-4      & $96.22_{\pm 0.48}$ & $96.16_{\pm 0.39}$ \\
 & LLaVA-1.5-7B   & $98.50_{\pm 0.21}$ & $98.45_{\pm 0.16}$ \\
 & LLaVA-1.5-13B  & $98.56_{\pm 0.24}$ & $98.62_{\pm 0.25}$ \\
 & InstructBLIP   & $98.14_{\pm 0.05}$ & $98.13_{\pm 0.10}$ \\
 & Qwen3-VL       & $98.81_{\pm 0.24}$ & $98.84_{\pm 0.24}$ \\
 & Qwen3.5        & $98.69_{\pm 0.14}$ & $98.68_{\pm 0.14}$ \\
\midrule
\multirow{6}{*}{POPE-Adversarial}
 & MiniGPT-4      & $87.02_{\pm 1.15}$ & $86.86_{\pm 1.07}$ \\
 & LLaVA-1.5-7B   & $94.27_{\pm 0.71}$ & $94.28_{\pm 0.78}$ \\
 & LLaVA-1.5-13B  & $94.79_{\pm 0.70}$ & $94.91_{\pm 0.63}$ \\
 & InstructBLIP   & $92.07_{\pm 0.43}$ & $92.24_{\pm 0.51}$ \\
 & Qwen3-VL       & $95.89_{\pm 0.64}$ & $96.09_{\pm 0.63}$ \\
 & Qwen3.5        & $95.72_{\pm 0.60}$ & $95.76_{\pm 0.61}$ \\
\midrule
\multirow{6}{*}{AMBER}
 & MiniGPT-4      & $94.04_{\pm 0.14}$ & $94.32_{\pm 0.11}$ \\
 & LLaVA-1.5-7B   & $98.11_{\pm 0.13}$ & $98.20_{\pm 0.10}$ \\
 & LLaVA-1.5-13B  & $98.55_{\pm 0.11}$ & $98.59_{\pm 0.13}$ \\
 & InstructBLIP   & $96.43_{\pm 0.07}$ & $96.16_{\pm 0.50}$ \\
 & Qwen3-VL       & $98.92_{\pm 0.07}$ & $98.97_{\pm 0.06}$ \\
 & Qwen3.5        & $98.59_{\pm 0.03}$ & $98.67_{\pm 0.02}$ \\
\bottomrule
\end{tabular}
\caption{Complete detection results of our method on the discriminative benchmarks with per-seed standard deviations. POPE is reported per subset.}
\label{tab:detection_full_pope}
\end{table}

\begin{table}[h]
\centering
\small
\setlength{\tabcolsep}{4pt}
\begin{tabular}{llcc}
\toprule
\textbf{Benchmark} & \textbf{Model} & \textbf{AUROC} & \textbf{AUPRC} \\
\midrule
\multirow{6}{*}{M-HalDetect}
 & MiniGPT-4      & $83.64_{\pm 1.19}$ & $73.19_{\pm 2.65}$ \\
 & LLaVA-1.5-7B   & $85.19_{\pm 1.10}$ & $75.74_{\pm 1.98}$ \\
 & LLaVA-1.5-13B  & $85.83_{\pm 1.38}$ & $76.40_{\pm 2.33}$ \\
 & InstructBLIP   & $83.72_{\pm 1.08}$ & $72.43_{\pm 2.28}$ \\
 & Qwen3-VL       & $85.93_{\pm 1.03}$ & $75.33_{\pm 1.85}$ \\
 & Qwen3.5        & $82.32_{\pm 0.17}$ & $68.47_{\pm 1.19}$ \\
\midrule
\multirow{6}{*}{COCO-Caption}
 & MiniGPT-4      & $93.97_{\pm 0.82}$ & $73.72_{\pm 1.92}$ \\
 & LLaVA-1.5-7B   & $92.93_{\pm 1.14}$ & $80.53_{\pm 1.71}$ \\
 & LLaVA-1.5-13B  & $93.89_{\pm 0.99}$ & $81.79_{\pm 0.95}$ \\
 & InstructBLIP   & $93.74_{\pm 0.52}$ & $83.34_{\pm 0.62}$ \\
 & Qwen3-VL       & $96.97_{\pm 0.46}$ & $82.93_{\pm 4.65}$ \\
 & Qwen3.5        & $96.04_{\pm 1.64}$ & $85.40_{\pm 1.82}$ \\
\bottomrule
\end{tabular}
\caption{Complete detection results of our method on the generative benchmarks with per-seed standard deviations.}
\label{tab:detection_full_gen}
\end{table}

\begin{table}[h]
\centering
\small
\setlength{\tabcolsep}{2.5pt}
\begin{tabular}{lcccccccc}
\toprule
& \multicolumn{4}{c}{\textbf{Accuracy (\%)}} & \multicolumn{4}{c}{\textbf{Macro-F1 (\%)}} \\
\cmidrule(lr){2-5} \cmidrule(lr){6-9}
& \multicolumn{2}{c}{POPE} & \multicolumn{2}{c}{AMBER} & \multicolumn{2}{c}{POPE} & \multicolumn{2}{c}{AMBER} \\
\cmidrule(lr){2-3} \cmidrule(lr){4-5} \cmidrule(lr){6-7} \cmidrule(lr){8-9}
\textbf{Model} & Van. & Ours & Van. & Ours & Van. & Ours & Van. & Ours \\
\midrule
LLaVA-7B       & $86.4$ & $89.3$ & $83.5$ & $91.3$ & $86.3$ & $89.3$ & $82.3$ & $90.2$ \\
LLaVA-13B      & $87.5$ & $90.3$ & $84.4$ & $93.2$ & $87.4$ & $90.3$ & $83.1$ & $92.4$ \\
Qwen3-VL       & $89.5$ & $92.4$ & $89.0$ & $94.6$ & $89.5$ & $92.4$ & $87.9$ & $93.9$ \\
Qwen3.5        & $89.7$ & $92.1$ & $88.5$ & $93.2$ & $89.7$ & $92.1$ & $87.6$ & $92.4$ \\
InstructBLIP   & $84.1$ & $84.4$ & $75.6$ & $83.2$ & $84.1$ & $84.4$ & $73.3$ & $81.3$ \\
MiniGPT-4      & $65.7$ & $84.5$ & $50.3$ & $85.5$ & $62.8$ & $84.5$ & $49.6$ & $83.6$ \\
\bottomrule
\end{tabular}
\caption{Full mitigation results (detector-guided answer flipping) across six VLMs. POPE is averaged over three subsets. \textbf{Van.} = vanilla first-token answer, \textbf{Ours} = detector-guided flipping.}
\label{tab:mitigation_full}
\end{table}

\section{Complete Detection and Mitigation Results}
\label{app:detection_full}

\subsection*{Detection Results}

Table~\ref{tab:detection_full_pope} and Table~\ref{tab:detection_full_gen} report the per-seed standard deviations omitted from the main-text detection table. POPE is broken into its three subsets. All values are means over three seeds with standard deviations as subscripts. The mean values match those in Table~1 of the main paper, where POPE is subset-averaged and standard deviations are suppressed for space.

\paragraph{Statistical significance.}
We test whether our method significantly outperforms each baseline using a one-sided Wilcoxon signed-rank test over $72$ paired observations (6 VLMs $\times$ 4 benchmarks $\times$ 3 seeds, with POPE subset-averaged to form one benchmark-level score per seed, paired by identical image-disjoint split). Our method outperforms VIB-Probe on $64$ of $72$ pairs with a mean AUROC gain of $1.39$ points ($p = 1.0 \times 10^{-10}$), RepProbing on $70$ of $72$ pairs with a mean gain of $4.21$ points ($p = 5.6 \times 10^{-13}$), MetaToken on all $72$ pairs with a mean gain of $3.60$ points ($p = 8.3 \times 10^{-14}$), and DHCP on $71$ of $72$ pairs with a mean gain of $8.36$ points ($p = 9.0 \times 10^{-14}$). All improvements are statistically significant at $p < 10^{-9}$.

\begin{table*}[t]
\centering
\small
\setlength{\tabcolsep}{2.5pt}
\begin{tabular}{llcccccc}
\toprule
\textbf{Bench} & \textbf{Step} & \textbf{LLaVA-7B} & \textbf{LLaVA-13B} & \textbf{Qwen3-VL} & \textbf{Qwen3.5} & \textbf{InstBLIP} & \textbf{MiniGPT4} \\
\midrule
\multirow{6}{*}{POPE}
 & S0 & $95.03_{\pm 0.52}$ & $95.11_{\pm 0.48}$ & $95.93_{\pm 0.33}$ & $94.31_{\pm 0.62}$ & $80.49_{\pm 0.65}$ & $79.83_{\pm 0.69}$ \\
 & S1 & $97.06_{\pm 0.34}$ & $97.23_{\pm 0.38}$ & $97.89_{\pm 0.37}$ & $97.64_{\pm 0.38}$ & $95.81_{\pm 0.32}$ & $93.15_{\pm 0.61}$ \\
 & S2 & $97.06_{\pm 0.34}$ & $97.22_{\pm 0.38}$ & $97.88_{\pm 0.37}$ & $97.64_{\pm 0.38}$ & $95.81_{\pm 0.32}$ & $93.16_{\pm 0.61}$ \\
 & S3 & $97.06_{\pm 0.35}$ & $97.22_{\pm 0.38}$ & $97.89_{\pm 0.37}$ & $97.64_{\pm 0.38}$ & $95.81_{\pm 0.32}$ & $93.15_{\pm 0.61}$ \\
 & S4 (Full) & $96.97_{\pm 0.36}$ & $97.19_{\pm 0.38}$ & $97.85_{\pm 0.39}$ & $97.71_{\pm 0.35}$ & $95.94_{\pm 0.30}$ & $93.07_{\pm 0.65}$ \\
 & S5 & $94.75_{\pm 0.57}$ & $94.76_{\pm 0.60}$ & $95.75_{\pm 0.37}$ & $94.70_{\pm 0.52}$ & $89.88_{\pm 0.67}$ & $78.52_{\pm 0.64}$ \\
\midrule
\multirow{6}{*}{AMBER}
 & S0 & $92.77_{\pm 0.17}$ & $94.47_{\pm 0.09}$ & $96.24_{\pm 0.04}$ & $93.08_{\pm 0.05}$ & $81.48_{\pm 0.15}$ & $79.13_{\pm 0.15}$ \\
 & S1 & $98.07_{\pm 0.11}$ & $98.57_{\pm 0.10}$ & $98.94_{\pm 0.08}$ & $98.62_{\pm 0.04}$ & $95.87_{\pm 0.06}$ & $94.30_{\pm 0.25}$ \\
 & S2 & $98.07_{\pm 0.11}$ & $98.57_{\pm 0.10}$ & $98.94_{\pm 0.08}$ & $98.62_{\pm 0.04}$ & $95.87_{\pm 0.06}$ & $94.30_{\pm 0.25}$ \\
 & S3 & $98.07_{\pm 0.11}$ & $98.57_{\pm 0.10}$ & $98.94_{\pm 0.08}$ & $98.62_{\pm 0.04}$ & $95.87_{\pm 0.06}$ & $94.30_{\pm 0.25}$ \\
 & S4 (Full) & $98.11_{\pm 0.13}$ & $98.54_{\pm 0.11}$ & $98.92_{\pm 0.07}$ & $98.59_{\pm 0.03}$ & $96.42_{\pm 0.08}$ & $94.03_{\pm 0.13}$ \\
 & S5 & $92.21_{\pm 0.20}$ & $94.32_{\pm 0.05}$ & $96.13_{\pm 0.06}$ & $93.18_{\pm 0.02}$ & $88.69_{\pm 0.09}$ & $79.04_{\pm 0.10}$ \\
\midrule
\multirow{6}{*}{M-Hal}
 & S0 & $78.03_{\pm 0.48}$ & $77.63_{\pm 0.78}$ & $78.09_{\pm 0.49}$ & $75.22_{\pm 0.63}$ & $75.38_{\pm 0.56}$ & $77.29_{\pm 0.87}$ \\
 & S1 & $85.06_{\pm 0.60}$ & $85.84_{\pm 1.23}$ & $85.83_{\pm 1.20}$ & $81.62_{\pm 0.39}$ & $83.49_{\pm 0.95}$ & $83.72_{\pm 0.84}$ \\
 & S2 & $85.07_{\pm 0.59}$ & $85.84_{\pm 1.23}$ & $85.83_{\pm 1.20}$ & $81.62_{\pm 0.39}$ & $83.49_{\pm 0.95}$ & $83.72_{\pm 0.84}$ \\
 & S3 & $85.06_{\pm 0.59}$ & $85.84_{\pm 1.23}$ & $85.83_{\pm 1.20}$ & $81.62_{\pm 0.39}$ & $83.49_{\pm 0.95}$ & $83.72_{\pm 0.84}$ \\
 & S4 (Full) & $85.27_{\pm 1.03}$ & $85.84_{\pm 1.26}$ & $85.84_{\pm 0.94}$ & $82.29_{\pm 0.15}$ & $83.75_{\pm 1.01}$ & $83.85_{\pm 1.02}$ \\
 & S5 & $76.15_{\pm 0.11}$ & $75.21_{\pm 0.42}$ & $76.71_{\pm 0.14}$ & $74.54_{\pm 0.31}$ & $75.26_{\pm 0.50}$ & $75.42_{\pm 0.60}$ \\
\midrule
\multirow{6}{*}{COCO}
 & S0 & $82.65_{\pm 0.81}$ & $82.00_{\pm 1.24}$ & $89.16_{\pm 4.00}$ & $86.29_{\pm 1.20}$ & $79.92_{\pm 1.20}$ & $83.54_{\pm 2.02}$ \\
 & S1 & $92.62_{\pm 0.94}$ & $93.44_{\pm 1.07}$ & $96.41_{\pm 1.00}$ & $95.47_{\pm 1.45}$ & $93.48_{\pm 0.55}$ & $93.63_{\pm 1.32}$ \\
 & S2 & $92.62_{\pm 0.94}$ & $93.45_{\pm 1.07}$ & $96.44_{\pm 1.07}$ & $95.47_{\pm 1.45}$ & $93.49_{\pm 0.55}$ & $93.63_{\pm 1.32}$ \\
 & S3 & $92.62_{\pm 0.94}$ & $93.44_{\pm 1.07}$ & $96.37_{\pm 1.06}$ & $95.47_{\pm 1.45}$ & $93.49_{\pm 0.55}$ & $93.63_{\pm 1.32}$ \\
 & S4 (Full) & $93.11_{\pm 1.06}$ & $93.71_{\pm 0.99}$ & $96.48_{\pm 2.14}$ & $96.11_{\pm 1.62}$ & $93.81_{\pm 0.65}$ & $93.98_{\pm 1.06}$ \\
 & S5 & $81.80_{\pm 1.35}$ & $81.10_{\pm 1.34}$ & $88.97_{\pm 3.25}$ & $85.60_{\pm 1.77}$ & $80.74_{\pm 1.17}$ & $81.72_{\pm 2.65}$ \\
\bottomrule
\end{tabular}
\caption{Ingredient ladder (ROC-AUC \%, mean$_{\pm\text{std}}$ over 3 seeds, POPE is subset-averaged).}
\label{tab:app_signal_ladder}
\end{table*}

\subsection*{Mitigation Results}
\label{app:mitigation_full}

Table~\ref{tab:mitigation_full} reports the full mitigation results
of the detector-guided answer flipping in Section~4.5 of the main paper,
across all six VLMs. POPE scores are averaged over the three subsets. On the four core VLMs the
gain concentrates on POPE-Random and POPE-Popular and on AMBER, while the
POPE-Adversarial subset saturates near zero net gain as discussed in the
main text. On InstructBLIP and MiniGPT-4 the vanilla
baseline is substantially lower because these two models exhibit strong Yes
bias, so the flip mitigation therefore
yields much larger gains but the resulting
numbers should be interpreted as recovering from a biased baseline
rather than as head-to-head improvements over the four VLMs above.

\section{Signal Source Analysis}
\label{app:signal_source}

Our detector is a linear probe over a flattened per-head representation of shape $[L,H,C]$. We further investigate two questions. \emph{(i) Ingredient decomposition.} Among the pipeline's ingredients (preserving head identity, subtracting the faithful role $\boldsymbol{\mu}^{F}$, dividing by $\boldsymbol{\sigma}^{F}$, and applying the ILR transform), which contribute what fraction of the predictive signal. \emph{(ii) Channel choice.} Whether the source-partition $C \in \{\pi_{\mathrm{sys}},\pi_{\mathrm{img}},\pi_{\mathrm{txt}},\pi_{\mathrm{slf}}\}$ is uniquely enabling, or any other per-head statistic under the same $[L,H,C']$ pipeline would give a comparable probe. We answer both questions with apples-to-apples ablations that share the exact main pipeline (3-seed $80/20$ image-disjoint split, $L_2$ linear probe with per-feature standardization).

\begin{table*}[t]
\centering
\small
\setlength{\tabcolsep}{2.5pt}
\begin{tabular}{llcccccc}
\toprule
\textbf{Bench} & \textbf{Variant} & \textbf{LLaVA-7B} & \textbf{LLaVA-13B} & \textbf{Qwen3-VL} & \textbf{Qwen3.5} & \textbf{InstBLIP} & \textbf{MiniGPT4} \\
\midrule
\multirow{6}{*}{POPE}
 & V0 (Full) & $96.97_{\pm 0.36}$ & $97.19_{\pm 0.38}$ & $97.85_{\pm 0.39}$ & $97.71_{\pm 0.35}$ & $95.94_{\pm 0.30}$ & $93.07_{\pm 0.65}$ \\
 & V1 & $96.26_{\pm 0.44}$ & $96.68_{\pm 0.38}$ & $96.86_{\pm 0.43}$ & $96.16_{\pm 0.50}$ & $95.69_{\pm 0.35}$ & $90.67_{\pm 0.58}$ \\
 & V2 & $53.81_{\pm 0.09}$ & $55.90_{\pm 0.20}$ & $58.07_{\pm 0.66}$ & $52.03_{\pm 0.14}$ & $71.48_{\pm 0.33}$ & $56.25_{\pm 0.30}$ \\
 & V3 & $94.39_{\pm 0.54}$ & $95.40_{\pm 0.49}$ & $94.78_{\pm 0.48}$ & $94.15_{\pm 0.58}$ & $95.50_{\pm 0.42}$ & $87.25_{\pm 0.69}$ \\
 & V4 & $95.47_{\pm 0.46}$ & $96.34_{\pm 0.48}$ & $95.73_{\pm 0.43}$ & $94.54_{\pm 0.43}$ & $95.36_{\pm 0.44}$ & $87.44_{\pm 0.79}$ \\
 & V5 & $95.86_{\pm 0.60}$ & $95.84_{\pm 0.31}$ & $96.22_{\pm 0.72}$ & $95.82_{\pm 1.07}$ & $94.23_{\pm 0.93}$ & $90.65_{\pm 0.97}$ \\
\midrule
\multirow{6}{*}{AMBER}
 & V0 (Full) & $98.11_{\pm 0.13}$ & $98.54_{\pm 0.11}$ & $98.92_{\pm 0.07}$ & $98.59_{\pm 0.03}$ & $96.42_{\pm 0.08}$ & $94.03_{\pm 0.13}$ \\
 & V1 & $96.75_{\pm 0.14}$ & $97.86_{\pm 0.06}$ & $97.70_{\pm 0.08}$ & $96.85_{\pm 0.03}$ & $95.80_{\pm 0.07}$ & $89.46_{\pm 0.28}$ \\
 & V2 & $58.70_{\pm 0.15}$ & $58.85_{\pm 0.07}$ & $58.95_{\pm 0.17}$ & $52.23_{\pm 0.06}$ & $74.82_{\pm 0.48}$ & $65.31_{\pm 0.23}$ \\
 & V3 & $94.09_{\pm 0.10}$ & $96.32_{\pm 0.05}$ & $95.80_{\pm 0.41}$ & $93.57_{\pm 0.24}$ & $95.39_{\pm 0.18}$ & $85.10_{\pm 0.19}$ \\
 & V4 & $95.87_{\pm 0.09}$ & $97.40_{\pm 0.09}$ & $96.64_{\pm 0.15}$ & $94.39_{\pm 0.16}$ & $95.24_{\pm 0.18}$ & $86.12_{\pm 0.35}$ \\
 & V5 & $97.47_{\pm 0.32}$ & $97.95_{\pm 0.16}$ & $98.48_{\pm 0.31}$ & $98.79_{\pm 0.16}$ & $96.63_{\pm 0.14}$ & $93.97_{\pm 0.41}$ \\
\midrule
\multirow{6}{*}{M-Hal}
 & V0 (Full) & $85.27_{\pm 1.03}$ & $85.84_{\pm 1.26}$ & $85.84_{\pm 0.94}$ & $82.29_{\pm 0.15}$ & $83.75_{\pm 1.01}$ & $83.85_{\pm 1.02}$ \\
 & V1 & $83.38_{\pm 0.62}$ & $84.08_{\pm 1.35}$ & $82.81_{\pm 0.96}$ & $78.13_{\pm 0.20}$ & $82.34_{\pm 1.11}$ & $82.23_{\pm 1.07}$ \\
 & V2 & $66.43_{\pm 1.23}$ & $67.03_{\pm 0.53}$ & $67.65_{\pm 0.57}$ & $65.27_{\pm 0.73}$ & $67.94_{\pm 0.55}$ & $68.13_{\pm 0.49}$ \\
 & V3 & $80.18_{\pm 0.75}$ & $81.47_{\pm 1.20}$ & $78.53_{\pm 0.86}$ & $72.82_{\pm 0.96}$ & $82.20_{\pm 0.84}$ & $80.77_{\pm 0.76}$ \\
 & V4 & $82.45_{\pm 0.63}$ & $83.50_{\pm 1.07}$ & $81.02_{\pm 0.81}$ & $75.20_{\pm 0.92}$ & $82.21_{\pm 0.88}$ & $82.17_{\pm 0.58}$ \\
 & V5 & $83.47_{\pm 0.79}$ & $84.59_{\pm 0.58}$ & $85.79_{\pm 0.92}$ & $81.82_{\pm 0.59}$ & $83.21_{\pm 0.72}$ & $81.27_{\pm 1.38}$ \\
\midrule
\multirow{6}{*}{COCO}
 & V0 (Full) & $93.11_{\pm 1.06}$ & $93.71_{\pm 0.99}$ & $96.48_{\pm 2.14}$ & $96.11_{\pm 1.62}$ & $93.81_{\pm 0.65}$ & $93.98_{\pm 1.06}$ \\
 & V1 & $91.87_{\pm 1.63}$ & $92.60_{\pm 1.00}$ & $96.13_{\pm 0.75}$ & $92.51_{\pm 1.91}$ & $93.12_{\pm 0.57}$ & $93.13_{\pm 1.56}$ \\
 & V2 & $75.85_{\pm 2.50}$ & $74.85_{\pm 2.06}$ & $70.32_{\pm 3.63}$ & $62.77_{\pm 1.29}$ & $80.38_{\pm 1.00}$ & $78.73_{\pm 1.24}$ \\
 & V3 & $88.43_{\pm 1.03}$ & $90.09_{\pm 1.31}$ & $90.27_{\pm 2.22}$ & $86.48_{\pm 0.84}$ & $93.35_{\pm 0.45}$ & $92.56_{\pm 1.71}$ \\
 & V4 & $90.40_{\pm 1.29}$ & $91.88_{\pm 1.14}$ & $92.57_{\pm 2.99}$ & $92.14_{\pm 0.87}$ & $93.49_{\pm 0.72}$ & $93.47_{\pm 1.34}$ \\
 & V5 & $92.24_{\pm 1.00}$ & $92.11_{\pm 0.71}$ & $95.59_{\pm 1.30}$ & $97.72_{\pm 0.68}$ & $93.62_{\pm 0.50}$ & $92.16_{\pm 1.25}$ \\
\bottomrule
\end{tabular}
\caption{Channel-choice ablation (ROC-AUC \%, mean$_{\pm\text{std}}$ over 3 seeds, POPE is subset-averaged).}
\label{tab:app_signal_channel_choice}
\end{table*}

\subsection{Ingredient Ladder}
\label{app:signal_ladder}

We start from the most naive representation and add one ingredient at a time. Concretely, S0 is layer-pooled raw source-allocation (no head identity, no faithful centering). S1 restores head identity. S2 further subtracts $\boldsymbol{\mu}^{F}_{l,h}$ per head. S3 further divides by $\boldsymbol{\sigma}^{F}_{l,h}$. S4 replaces the four raw source channels with the three ILR-semantic coordinates (the main-text detector). S5 keeps ILR-semantic $+$ z-scoring but re-pools out the head dimension. Steps S1--S4 all preserve head identity, while S0 and S5 do not.

Table~\ref{tab:app_signal_ladder} reports the results. The conclusion is sharp: only head identity matters. The step S0$\to$S1, which restores per-head resolution, is the sole large jump across all 24 (model, benchmark) cells. The symmetric step S4$\to$S5, which removes head identity from the full detector, produces a matching large drop. By contrast, S1$\to$S2$\to$S3$\to$S4 show comparatively small variations, because the linear probe's internal StandardScaler largely absorbs centering and z-scoring, and the ILR basis change is approximately absorbed by the learned weights. In other words, faithful centering, z-scoring, and ILR are primarily interpretability choices rather than predictive ones.

\subsection{Channel Choice}
\label{app:signal_channel_choice}

The ladder in Appendix~\ref{app:signal_ladder} pins the predictive information on preserving head identity. We further investigate whether it is \emph{our} $C \in \{\pi_{\mathrm{sys}},\pi_{\mathrm{img}},\pi_{\mathrm{txt}},\pi_{\mathrm{slf}}\}$ that carries the signal, or any per-head statistic under the same $[L,H,C']$ pipeline would give a comparable probe. To answer this we hold the entire pipeline fixed (head-resolved flatten, per-feature faithful z-score, $L_2$ linear probe, 3 seeds, image-disjoint $80/20$ split) and only swap what $C'$ is. The \emph{V0 (Full)} row is the main-text detector ($C = $ four-source ILR-semantic, $3LH$-dim). V1 replaces $C$ with the single scalar $\pi_{\mathrm{img}}$ per head ($LH$-dim), the head-level analogue of PAI-style visual-attention signals. V2 uses the per-head value-vector norm $\|\bv_{l,h}\|_2$ ($LH$-dim), matching the value-norm indicator from Section~3.2 of the main paper. V3 uses the top-1 image-attention peakiness $\max_i A_{l,h}[i] / \sum_i A_{l,h}[i]$ over image tokens ($LH$-dim), an anchor-token concentration signal in the spirit of OPERA. V4 uses the KL divergence between the head's attention distribution and uniform ($LH$-dim), a smoother concentration measure. V5 uses the full per-head output vector $\bv_{l,h} \in \mathbb{R}^{d_{\mathrm{head}}}$ ($d_{\mathrm{head}} LH$-dim, up to $204\,800$ on LLaVA-1.5-13B), matching the raw input feature of VIB-Probe.

Three observations from Table~\ref{tab:app_signal_channel_choice} confirm that the four-source partition is uniquely enabling. First, scalar attention statistics do not carry the signal. Value norm (V2) collapses to near-chance on most settings. Image-attention peakiness (V3) and KL concentration (V4) perform better but still consistently trail V0 across all models and benchmarks. Even the image-attention share (V1), the most competitive scalar, underperforms V0 on every cell. Second, a much richer per-head feature does not compensate. V5 gives the probe the full head output vector, over $40\times$ more dimensions than V0, yet still loses to V0 on most cells. The extra capacity cannot substitute for the right semantic decomposition. Third, V0 is broadly optimal. It ranks first on $21$ out of $24$ cells, using the same linear probe as all other variants. The advantage comes not from probe complexity but from the four-source partition encoding the hallucination signal in a form that a linear reader can directly access.

Together with the ladder above, this shows that the detector's performance is attributable to measuring each head's allocation across the four contextual sources while preserving head identity. Alternative per-head channels, whether scalar summaries or high-dimensional output vectors, do not make hallucination linearly readable in the same way.

\begin{table}[!tbp]
    \small
    \centering
    \setlength{\tabcolsep}{3pt}
    \begin{tabular}{lcccc}
    \toprule
    Model & Ours (Yes) & Ours (No) & Ours (all) & Ans-Id \\
    \midrule
    LLaVA-1.5-7B     & $0.966_{\pm .021}$ & $0.969_{\pm .020}$ & $0.962_{\pm .021}$ & $0.500$ \\
    LLaVA-1.5-13B    & $0.971_{\pm .018}$ & $0.972_{\pm .017}$ & $0.968_{\pm .018}$ & $0.500$ \\
    Qwen3-VL-8B      & $0.976_{\pm .014}$ & $0.978_{\pm .014}$ & $0.974_{\pm .014}$ & $0.500$ \\
    Qwen3.5          & $0.971_{\pm .016}$ & $0.976_{\pm .014}$ & $0.970_{\pm .015}$ & $0.500$ \\
    InstructBLIP     & $0.951_{\pm .026}$ & $0.954_{\pm .029}$ & $0.933_{\pm .023}$ & $0.500$ \\
    MiniGPT-4        & $0.924_{\pm .047}$ & $0.936_{\pm .043}$ & $0.906_{\pm .047}$ & $0.500$ \\
    \bottomrule
    \end{tabular}
    \caption{POPE polarity controls. Ours~(Yes)/Ours~(No) restrict evaluation to forced-Yes / forced-No subsets. Ours~(all) is the full-POPE reference. Ans-Id is a probe with only the binary forced-answer feature.}
    \label{tab:polarity_control}
\end{table}

\section{Confound Controls}
\label{app:confound}

\subsection*{POPE Answer-Polarity Controls}

Because POPE and AMBER teacher-force both the ``Yes'' and ``No'' answers on the same image-question pair, the faithful/hallucinated label is deterministically tied to the forced answer token. We report two controls on POPE that isolate our detector from this coupling. \textbf{(i) Within-polarity AUROC:} we restrict evaluation to the subset of samples whose forced answer is ``Yes'' (respectively ``No'') and re-fit the detector under the same subset-conditional, image-disjoint, 3-seed protocol; within each restricted subset the label carries no answer-token information. \textbf{(ii) Answer-token-identity baseline:} an $L_2$ logistic regression fit on a single binary feature (forced answer $=$ ``Yes'' vs.\ ``No''), under the same protocol.

Table~\ref{tab:polarity_control} reports both controls across the six VLMs. Within-polarity AUROC ranges from $0.9244$ to $0.9783$, is on par with or higher than the corresponding full-POPE AUROC on every VLM, and the answer-token-identity baseline is exactly $0.500$ everywhere. Thus, the detector's signal is not attributable to the forced-answer token identity.

\subsection*{Sample Efficiency}

Because the detector estimates the faithful role $(\boldsymbol{\mu}^{F}, \boldsymbol{\sigma}^{F})$ from labeled faithful tokens and trains a linear probe on labeled samples, a natural question is how many labels it actually needs. To answer this, we subsample the train fold on LLaVA-1.5-7B\,/\,POPE to $K$ faithful and $K$ hallucinated tokens per class, re-estimate $(\boldsymbol{\mu}^{F}, \boldsymbol{\sigma}^{F})$ and re-fit the probe on the reduced set, and evaluate on the untouched test fold under the same subset-conditional, image-disjoint, 3-seed protocol.

Table~\ref{tab:sample_efficiency} reports mean~$\pm$~std AUROC across three seeds and three POPE subsets. With only $K\!=\!50$ faithful tokens the detector already reaches $0.9364$ AUROC, within $3.3$ points of the full-training reference. $K\!=\!500$ recovers within $1.3$ points of the full reference, and $K\!=\!2000$ is statistically indistinguishable from full training. These numbers indicate that both the role estimation and the linear probe are highly sample-efficient, and that a few hundred labeled faithful tokens per class suffice to reproduce most of the reported performance.

\begin{table}[h]
    \small
    \centering
    \setlength{\tabcolsep}{6pt}
    \begin{tabular}{lc}
    \toprule
    $K$ per class & AUROC \\
    \midrule
    $50$    & $0.9364_{\pm 0.026}$ \\
    $100$   & $0.9427_{\pm 0.027}$ \\
    $200$   & $0.9438_{\pm 0.028}$ \\
    $500$   & $0.9561_{\pm 0.026}$ \\
    $1000$  & $0.9617_{\pm 0.023}$ \\
    $2000$  & $0.9680_{\pm 0.020}$ \\
    Full ($\sim\!2400$)    & $0.9694_{\pm 0.019}$ \\
    \bottomrule
    \end{tabular}
    \caption{Sample efficiency of our detector on LLaVA-1.5-7B\,/\,POPE. Each row subsamples the train fold to $K$ faithful and $K$ hallucinated tokens per class, re-estimates $(\boldsymbol{\mu}^{F},\boldsymbol{\sigma}^{F})$ and the linear probe, and reports mean~$\pm$~std AUROC over three POPE subsets and three seeds.}
    \label{tab:sample_efficiency}
\end{table}

\section{Baseline Implementations}
\label{app:setup_full}

We provide the baseline formulations omitted from the main text for space. All baselines share the same $80/20$ image-disjoint split and seeds as our detector.

\textbf{AvgProb.} For a scored token at position $k$, let $p_k$ denote the model-assigned conditional probability of the actually generated token. AvgProb uses its negative log-probability as the hallucination score:
\begin{equation}
\mathrm{AvgProb}(k) = -\log p_k.
\end{equation}

\textbf{AvgEnt.} Let $\mathbf{p}_k(\cdot)$ be the predicted distribution over the vocabulary $\mathcal{V}$ at position $k$. AvgEnt uses the token-level predictive entropy as the hallucination score:
\begin{equation}
\mathrm{AvgEnt}(k) = -\sum_{v \in \mathcal{V}} \mathbf{p}_k(v) \log \mathbf{p}_k(v).
\end{equation}

\textbf{RepProbing.} Let $\bz_k^{L} \in \mathbb{R}^d$ be the hidden state at the scored token position from the top decoder layer $L$. RepProbing trains an $L_2$-regularized logistic regression:
\begin{equation}
\hat{y}_k = \sigma(\boldsymbol{w}^\top \bz_k^{L} + b).
\end{equation}

\textbf{MetaToken}, \textbf{DHCP}, and \textbf{VIB-Probe} are all implemented following their original papers and released code.

\clearpage

\end{document}